\documentclass[10pt,twocolumn,letterpaper]{article}

\usepackage[pagenumbers]{cvpr} 

\usepackage[dvipsnames]{xcolor}

\definecolor{cvprblue}{rgb}{0.21,0.49,0.74}
\usepackage[pagebackref,breaklinks,colorlinks,citecolor=cvprblue]{hyperref}
\usepackage{tabularx}
\usepackage{multirow}
\usepackage{stmaryrd}
\DeclareMathOperator*{\argminA}{arg\,min} 
\DeclareMathOperator*{\minA}{min} 
\def\paperID{*****} 
\def\confName{CVPR}
\def\confYear{2024}
\title{Large-scale DSM registration via motion averaging}

\author{Ningli Xu\\
The Ohio State University\\
Columbus, Ohio, USA\\
{\tt\small xu.3961@buckeyemail.osu.edu}
\and
Rongjun Qin*\\
The Ohio State University\\
Columbus, Ohio, USA\\
{\tt\small qin.324@osu.edu}
}

\begin{document}
\maketitle
\begin{abstract}
Generating wide-area digital surface models (DSMs) requires registering a large number of individual, and partially overlapped DSMs. This presents a challenging problem for a typical registration algorithm, since when a large number of observations from these multiple DSMs are considered, it may easily cause memory overflow. Sequential registration algorithms, although can significantly reduce the computation, are especially vulnerable for small overlapped pairs, leading to a large error accumulation. In this work, we propose a novel solution that builds the DSM registration task as a motion averaging problem: pair-wise DSMs are registered to build a scene graph, with edges representing relative poses between DSMs. Specifically, based on the grid structure of the large DSM, the pair-wise registration is performed using a novel nearest neighbor search method. We show that the scene graph can be optimized via an extremely fast motion average algorithm with O(N) complexity (N refers to the number of images). Evaluation of high-resolution satellite-derived DSM demonstrates significant improvement in computation and accuracy.

\textbf{This article has been accepted for publication in ISPRS Annals of the Photogrammetry, Remote Sensing and Spatial Information Sciences, Volume X-1-2024 ISPRS TC 1 Mid-term Symposium, 13-17 May 2024, Changsha, China}
\end{abstract}    
\section{Introduction}
\label{sec:intro}

Generating wide-area digital surface models (DSMs) is pivotal in establishing foundational geospatial datasets. This often involves registering a large number of individual DSMs generated either from the same or different sources, such as from satellite stereo-based reconstruction (e.g. WorldView 2 \& 3\footnote{https://resources.maxar.com/data-sheets/worldview-2}, PlanetScope \cite{huang2022evaluation,qin2023using}), aerial sensors \cite{xu2024multi,huang2023critical}, lidar, Synthetic Aperture Radar (SAR) (e.g. SRTM \cite{srtm}).  These individual DSMs, however, may have varying degrees of absolute geometric accuracy, as well as partial and even very minimal overlap. Thus, registering DSMs under this context presents as a challenging problem both due to both the scale of the problem, as well as the sub-optimal quality of input. For example, registering these DSMs may require minimizing geometric errors between billions of points, easily causing memory overflow, at the same time, registering partial and low-overlap DSMs easily brings large errors leading to error accumulation when multiple DSMs are registered. To this end, We propose a framework based on motion averaging, which entails enumerating all overlapped DSM pairs to establish a scene graph. Our approach involves the utilization of an efficient pair-wise registration method to eliminate systematic errors at the pair-wise level and subsequently redistribute these errors across the graph. This process aims to reduce and evenly distribute systematic errors among multiple DSMs.


As mentioned above, the task of registering multiple DSMs raises several unique and difficult challenges. To be more specific, first of all, the huge volume of data adds to the difficulty of memory and computation. The standard method is the iterative closest point (ICP) \cite{besl1992method}, which iteratively minimizes the cloud-to-cloud distance between a pair of 3D data. The correspondences are formed by searching for the nearest neighbor (NN) point of reference data for each query point, which usually can be sped up by k-d tree \cite{besl1992method,greenspan2003approximate}, or octree \cite{eggert2012octree}, but the initialization of a k-d tree necessitates caching the entire reference data into memory, requiring \(O(N)\) space, \(O(NlogN)\) time. Such methods become impractical for city or terrain-level data. For instance, caching a single WorldView-2 DSM (usually around \(20000 \times 20000\) equals 400 million points) consumes 22GB for the construction of k-d tree. Another challenge is accurately estimating global poses from pair-wise registration results. The prevailing method adopts a greedy approach, initiating with the initial pair and estimating subsequent DSMs based on maximum overlap. However, as the DSM count increases, this approach accumulates small registration errors, resulting in a significant deviation from the actual scene.

\begin{figure*}[tb]
    \centering
        \includegraphics[width=2\columnwidth]{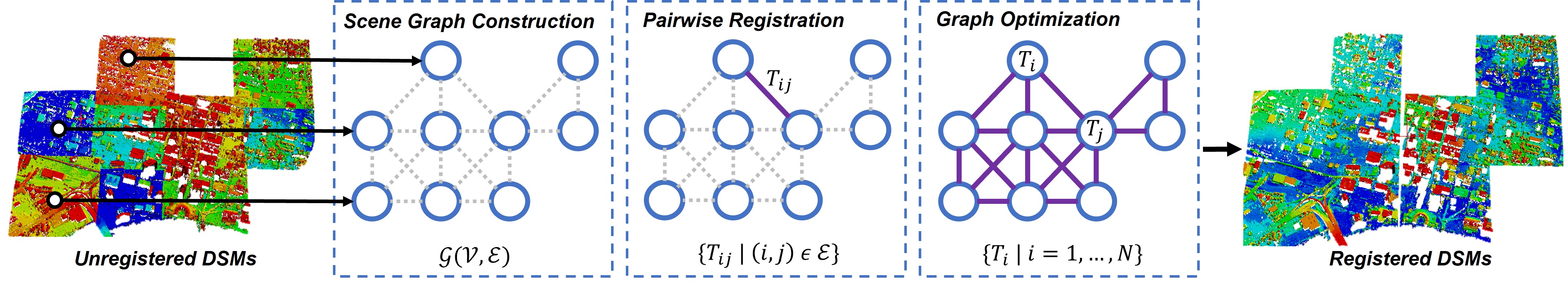}
    \caption{Overview of the proposed method. Given \(N\) unaligned DSMs, our target is to remove their systematic errors to produce seamless registered DSMs (the unaligned and registered DSMs are color-coded based on their heights). We first construct a scene graph \(\mathcal{G}(\mathcal{V},\mathcal{E})\) with edges denoting all possible pairs of DSMs.Then, the proposed DSM-ICP (see \autoref{sec:31}) is performed for all edges determining their pair-wise transformation \(\{T_{ij} | (i,j) \in \mathcal{E}\}\). Finally, a motion averaging approach is performed to estimate the global poses \(\{T_i | i=1,...,N\}\) (see \autoref{sec:32})}
    \label{fig:pipeline}
\end{figure*}

To address the above-mentioned challenges, we propose a motion averaging-based framework for removing the systematic errors among multiple unaligned DSMs, as shown in \autoref{fig:pipeline}. We first propose an ICP-based pair-wise registration method that can perform fast and exact NN searching by utilizing the grid structure of DSM (termed as DSM-ICP). Its space and computation complexity are independent of the data volume, allowing excellent scalability to large-scale datasets. Based on it, we perform pair-wise registration to all possible pairs of DSM, establishing a reliable scene graph, where each edge represents the relative poses and carries weights based on its overlap and registration error. Finally, a motion averaging approach over the weighted edge graph is performed to redistribute the registration error for the estimation of the global poses. In our experiments, we quantitatively and qualitatively evaluate our proposed method using a large number of individual satellite-stereo-based DSMs against the ground-truth lidar point cloud data.
The rest of this paper is organized as follows: \autoref{sec:related_work} reviews the related works of 3D registration methods. \autoref{sec:methodology} introduces the proposed motion averaging framework in detail. \autoref{sec:experiment} presents the experimental results, comparative studies, and our analysis. \autoref{sec:conclusion} concludes this work. 

\section{Related Work}\label{sec:related_work}
\textbf{Pair-wise 3D registration} is the process of estimating the transformation (rigid-body transformation in our case) between a pair of 3D data sets to minimize the systematic error between two 3D data, which is widely used in 3D mapping and change detection \cite{xu2021volumetric} tasks. It can be categorized into coarse and fine registration according to the registration accuracy. Coarse registration assumes no initial alignment, involving the extraction of key primitives (e.g. key points \cite{rusu2009fast,ao2021spinnet}, lines \cite{chen2019feature}, planes \cite{chen2019plade}, 4-points congruent set \cite{mellado2014super}), correspondence construction \cite{bai2021pointdsc,zhang20233d}, and transformation estimation \cite{barath2018graph,yang2020teaser}. Such methods often yield suboptimal accuracy due to errors in primitive localization and matching, necessitating subsequent fine registration \cite{xu2023point}. The standard fine registration method is the ICP algorithm, which iteratively refines the transformation by searching for NN points and estimating the transformation. Variants of ICP aim to enhance its robustness (Sparse-ICP \cite{bouaziz2013sparse}, Robust ICP \cite{zhang2021fast}) and convergence speed (Point-to-Plane ICP \cite{arun1987least}, Symmetric ICP \cite{rusinkiewicz2019symmetric}, AA-ICP \cite{pavlov2018aa}). The NN search typically employs k-d tree, requiring \(O(N)\) memory for reference data caching and \(O(logN)\) time for querying. Typically, CODEM\footnote{https://github.com/NCALM-UH/CODEM} is also an ICP-based method for DSM registration while they transform DSM into point cloud format and perform standard ICP. In comparison, ours utilizes the grid structure of DSM and aims to optimize the memory and computation requirements of NN search to make it applicable to large-scale datasets.

\textbf{Multi-view registration} considers each DSM as a view, and simultaneously solves the registration of each DSM at the same time. This can be accomplished through growing-based and optimization-based approaches. Growing-based methods initiate registration with the pair possessing the largest overlap and progressively register new data to existing data. The order of data registration is determined using Kruskal's algorithm \cite{kruskal1956shortest}. This type of method is efficient while as the number of data sets increases, the accumulated registration error will lead to drift problems. Optimization-based methods involve constructing a scene graph, with each edge denoting relative pose and global pose estimation achieved by minimizing a predefined objective function — a concept known as motion averaging or pose graph optimization in the robotics community \cite{carlone2015initialization}. The closed-form solutions estimate the rotation using the singular value decomposition (SVD)-based methods \cite{arie2012global,gojcic2020learning,wang2023robust} and then solve the translation using the least square methods \cite{gojcic2020learning}. The iterative least-squares methods linearize the objective function and use Gauss-Newton or Levenberg-Marquardt methods to update the global poses. Our solution involves the SVD-based method to ensure computation and memory efficiency.

\section{METHODOLOGY}\label{sec:methodology}
Given a set of unaligned DSMs \(\mathcal{P}=\{P_i|i=1,...,N\}\), our goal is to recover the global poses \(\{T_i=(R_i,t_i) \in SE(3) | i=1,...,N\}\). In the following, we first introduce a computation and memory-efficient DSM registration method in \autoref{sec:31}. Based on it, we construct a scene graph with reliable weights for each edge and apply a motion-averaging approach to estimate the poses of every DSM in \autoref{sec:32}.

\subsection{DSM-ICP}\label{sec:31}
Aligning a pair of DSMs \(P\) and \(Q\) involves finding the rigid transformation \((R|t)\) such that applying the transformation to \(P\) causes the two DSMs to be as close as possible. The standard method is iterative closet point (ICP). It minimizes the distance between two DSMs by alternating between two steps:
\begin{itemize}
    \item \textit{Correspondence step}: search pairs of corresponding points \((p_i,q_i^k)\), where \(q_i^k\) is the closest point to \(p_i\) given the current transformation \((R^k|t^k)\):
\begin{equation}
    q_i^k=\argminA_{q \in Q} \|R^kp_i+t^k-q\|
\end{equation}
    \item \textit{Estimation step:} estimate the new transformation \((R^{k+1},t^{k+1})\) by optimizing the \(\ell_2\) distance given the current correspondences \((p_i,q_i^k)\):
\begin{equation}\label{eq2}
    (R^{k+1},t^{k+1})=\argminA_{R,t}\sum_{i} {\|Rp_i+t-q_i^k\|}^2
\end{equation}
\end{itemize}

Common methods solve the correspondence step by applying NN methods, such as k-d tree \cite{besl1992method}, approximate k-d tree \cite{greenspan2003approximate}, or octree \cite{eggert2012octree}. These methods necessitate the initialization of the spatial structures to cache the whole reference data \(Q\).Specifically, k-d tree needs \(O(N)\) memory and \(O(NlogN)\) time, and each NN searching requires \(O(logN)\) time. In the estimation step, the closed-form solution involves separating the rotation and translation. Rotation estimation is achieved through SVD-based methods, while the translation can be readily determined.

\textbf{Efficient and exact NN searching.} In a DSM, each pixel denotes a 3D point in world coordinates, where its uv coordinates \([u_i,v_i]\) indicate the geo-location in the horizontal plane and its value \(h_i=dsm[u_i,v_i]\) represents the height in meters. The neighboring pixels within a range of \(x\) pixels can be queried in constant time using the operation \([u_i-x:u_i+x,v_i-x:v_i+x]\). Given a query pixel \([u_p,v_p]\) from the moving DSM \(P\), we will introduce an efficient method to find the upper bound of its NN from the reference DSM \(Q\) in constant time.

\begin{figure}[tb]
    \centering
        \subcaptionbox{Initial phase}{\includegraphics[width=0.49\columnwidth]{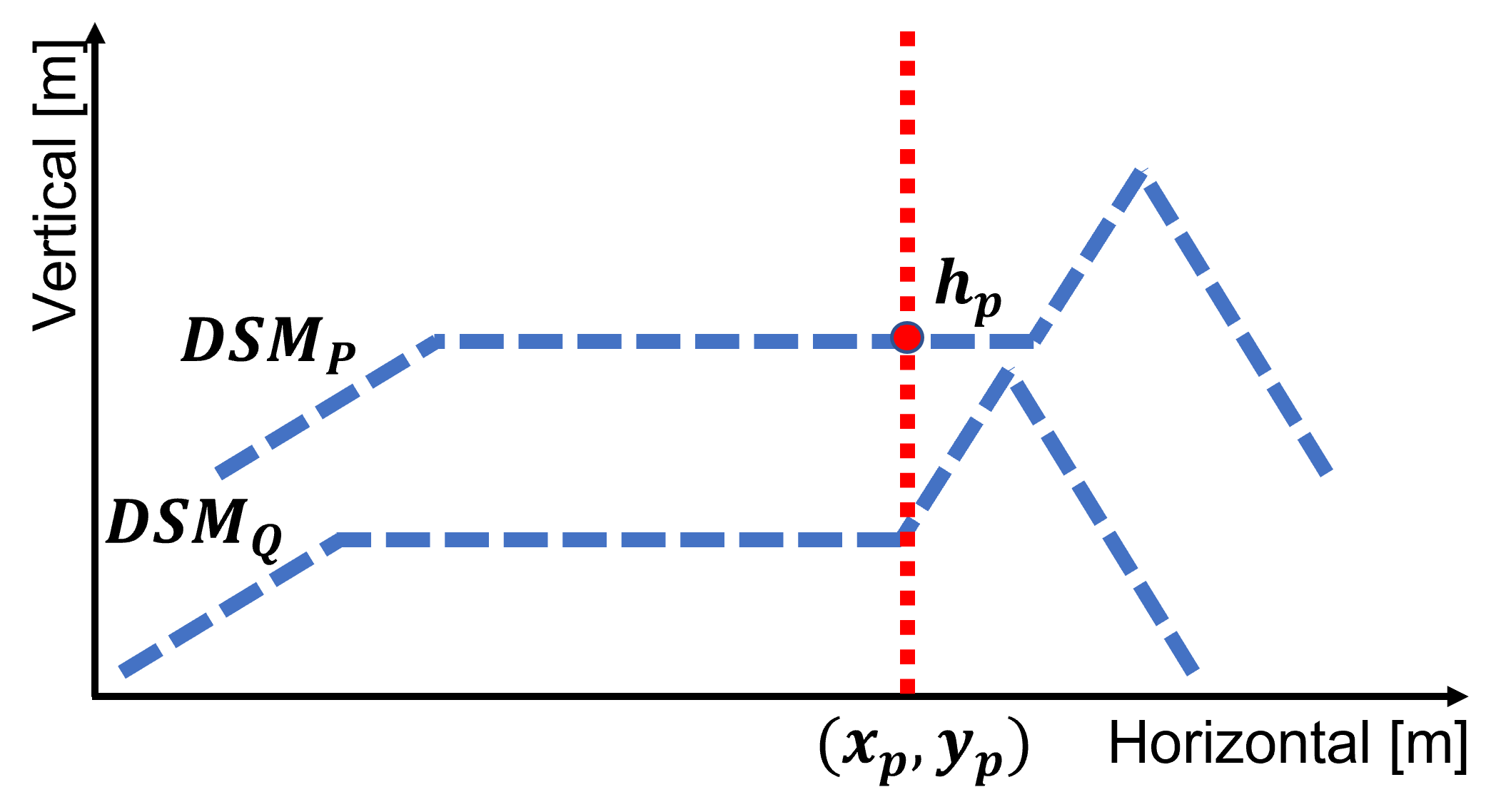}}
        \subcaptionbox{Correspondence phase}{\includegraphics[width=0.49\columnwidth]{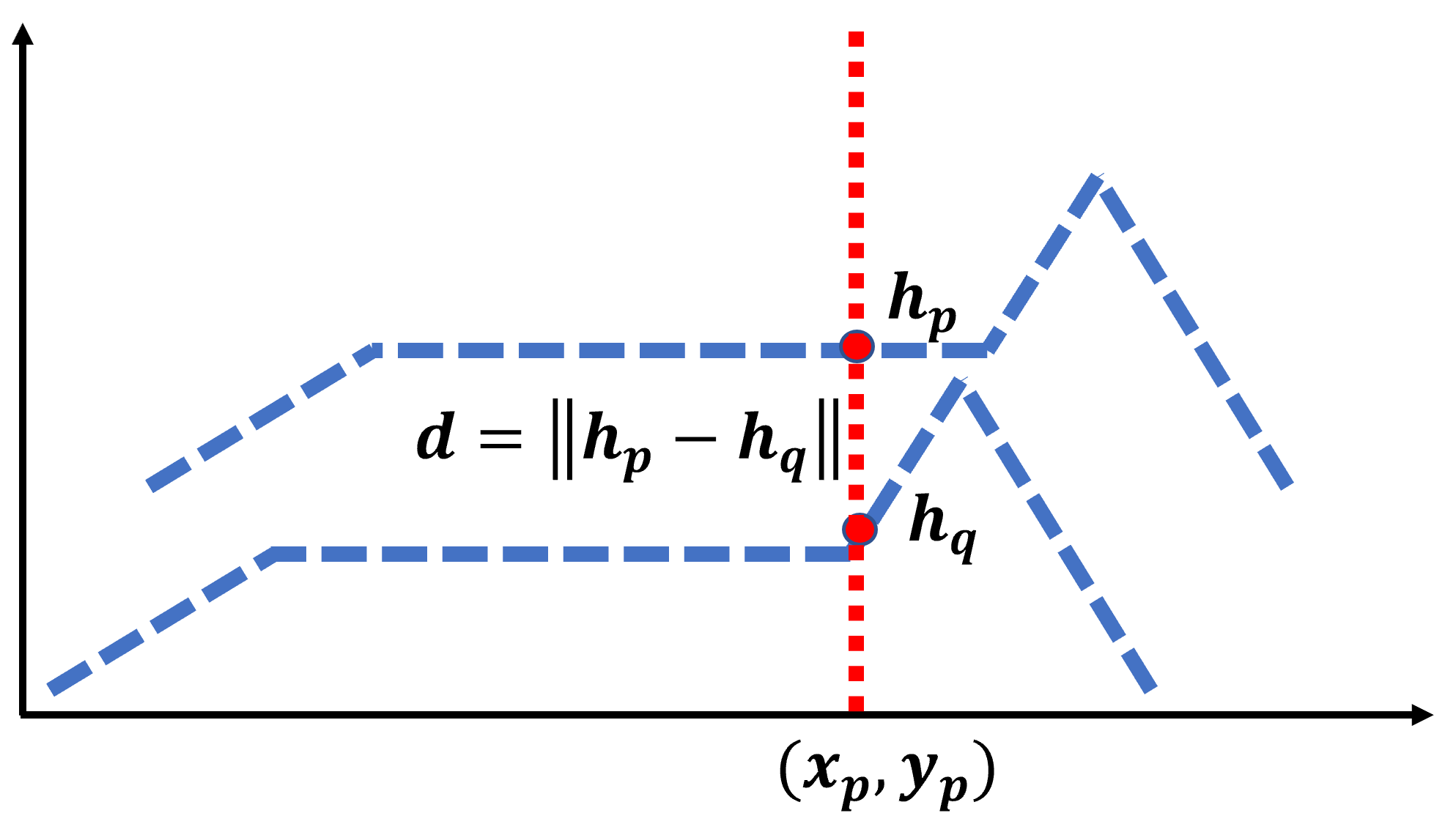}}
        \subcaptionbox{Bounding phase}{\includegraphics[width=0.49\columnwidth]{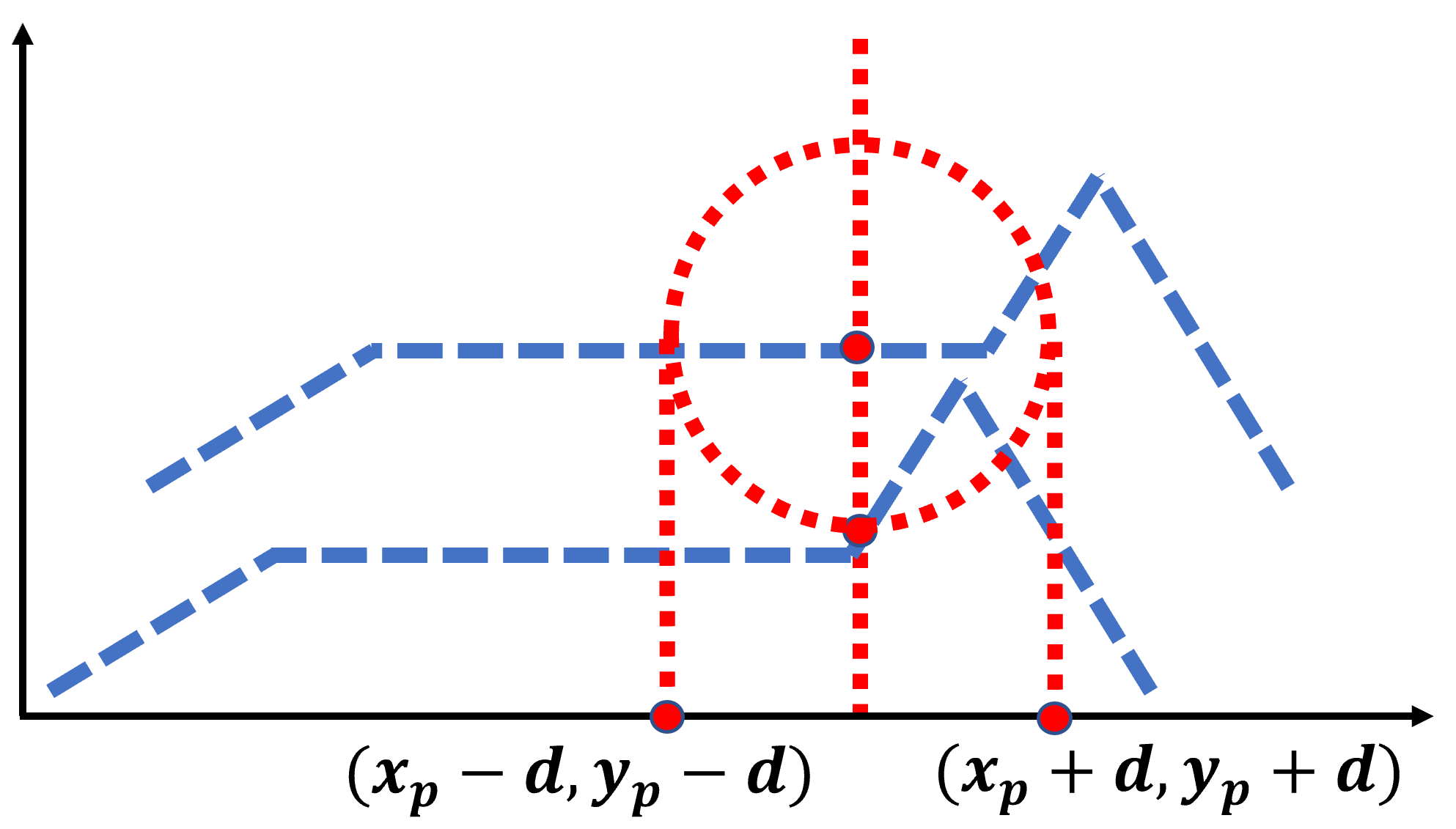}}
        \subcaptionbox{NN phase}{\includegraphics[width=0.49\columnwidth]{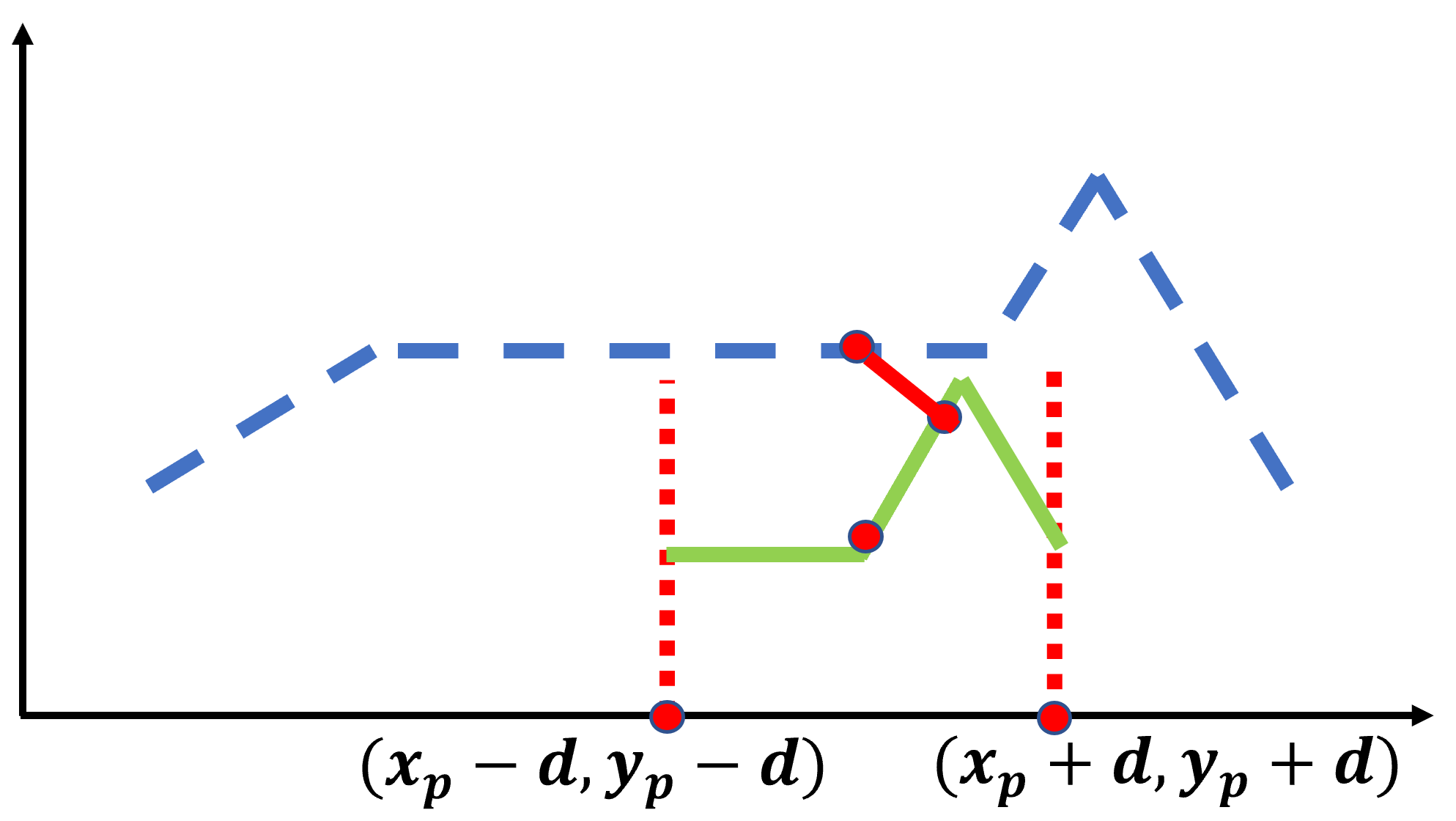}}
    \caption{Illustration of the proposed NN search method. To simplify, DSMs are depicted as profiles in 2D space, with the x-axis representing the horizontal plane and the y-axis denoting height. The "Blue dashed" lines depict DSM data stored on disk, while the "Green solid" line represents cached data in RAM.}
    \label{fig:dsmicp}
\end{figure}

Specifically, at the initial phase shown in \autoref{fig:dsmicp}, we calculate its world coordinate \((x_p,y_p)\) using \autoref{eq:eq3}. Then, the pixel coordinate \((u_q,v_q)\) from the other DSM \(Q\) covering the same world coordinates can be calculated using \autoref{eq:eq3}. The height of that pixel can be looked up and termed as \(h_q=DSM_q(u_q,v_q)\). 

\begin{equation}
    \begin{split}
            (x_p,y_p,h_p)=uv2world(u_p,v_p,tfw_p)\\
    (u_q,v_q)=world2uv(x_p,y_p,h_p,tfw_q)
    \end{split}
\end{equation} \label{eq:eq3}

Then, the pixel \((u_q,v_q)\) serves as the initial correspondence to the query pixel \((u_p,v_p)\), which is found in constant time. Moreover, their Euclidean distance also serves as the upper bound that the NN is guaranteed to be located within the 3D sphere ball centered at the query 3D point with a radius of the \(d\), where \(d=\|h_p-h_q\|\).

At the bounding phase, the 3D sphere ball is projected into the horizontal plane, formulating the 2D rectangle \([x_q-d:x_q+d,y_q-d:y_q+d]\) as the final bound. Finally, at the NN phase, the uv coordinates of the bounding rectangle are calculated and any pixels within the bounding rectangle are cached. By traversing all the cached pixels, we can find the NN.

\begin{figure}[tb]
    \centering
        \subcaptionbox{JAX2}{\includegraphics[width=0.3\columnwidth]{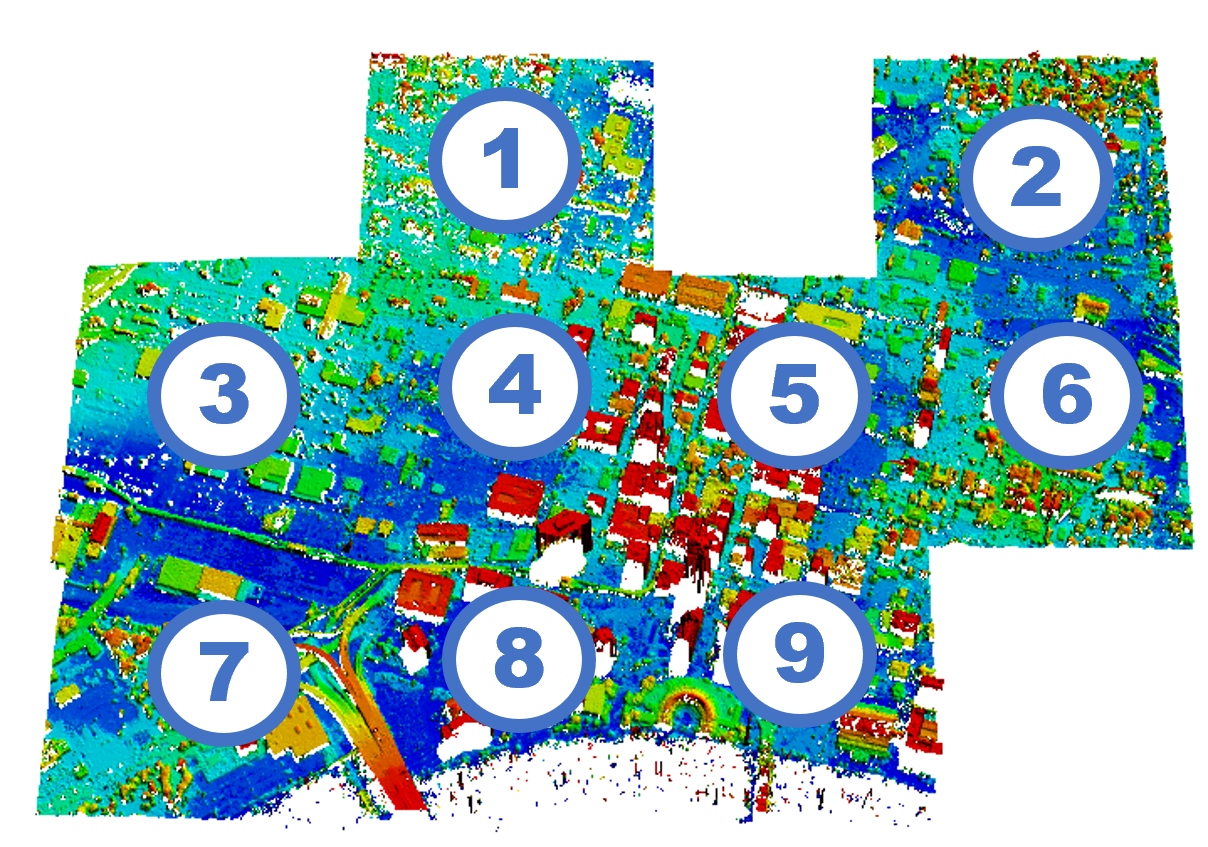}}
        \subcaptionbox{JAX2 : MST graph}{\includegraphics[width=0.3\columnwidth]{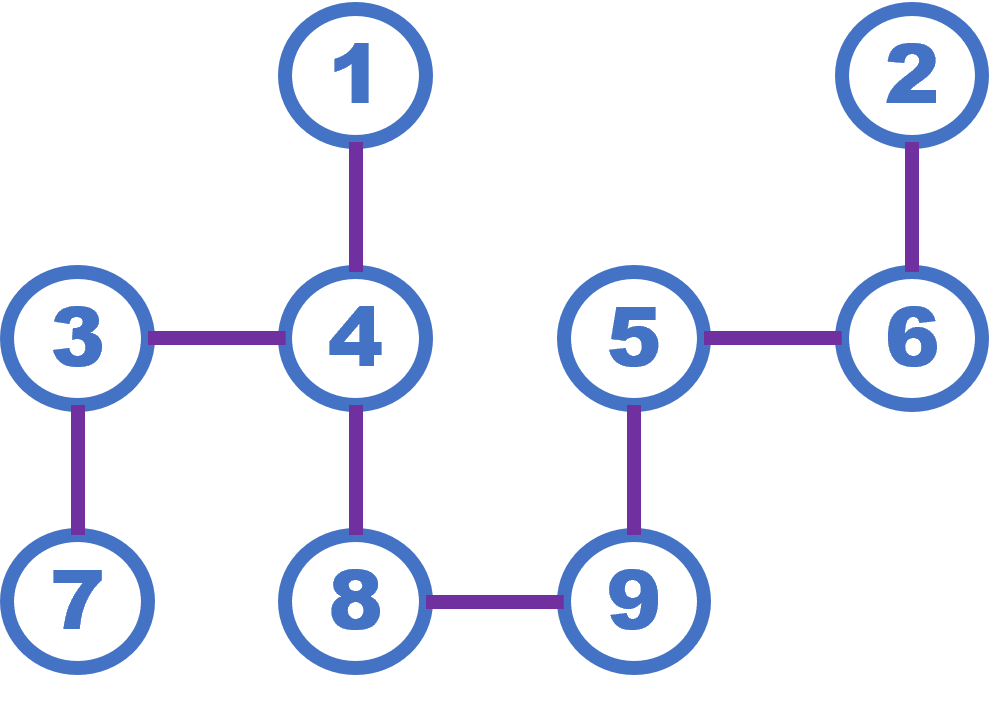}}
        \subcaptionbox{JAX2 : our graph}{\includegraphics[width=0.3\columnwidth]{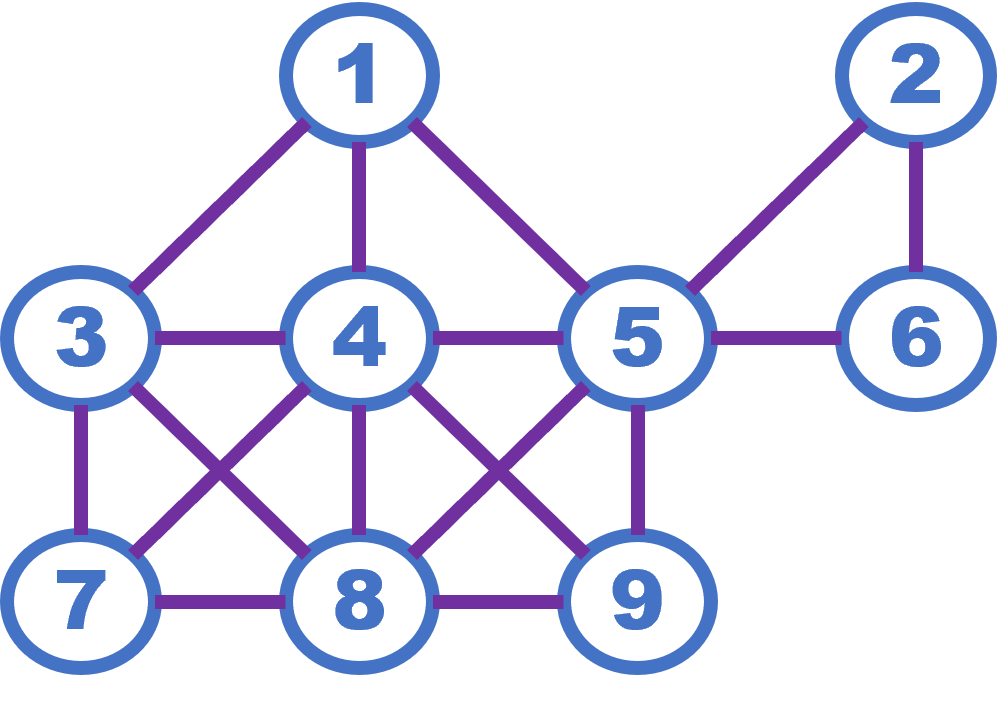}}
        \subcaptionbox{OMA3}{\includegraphics[width=0.25\columnwidth]{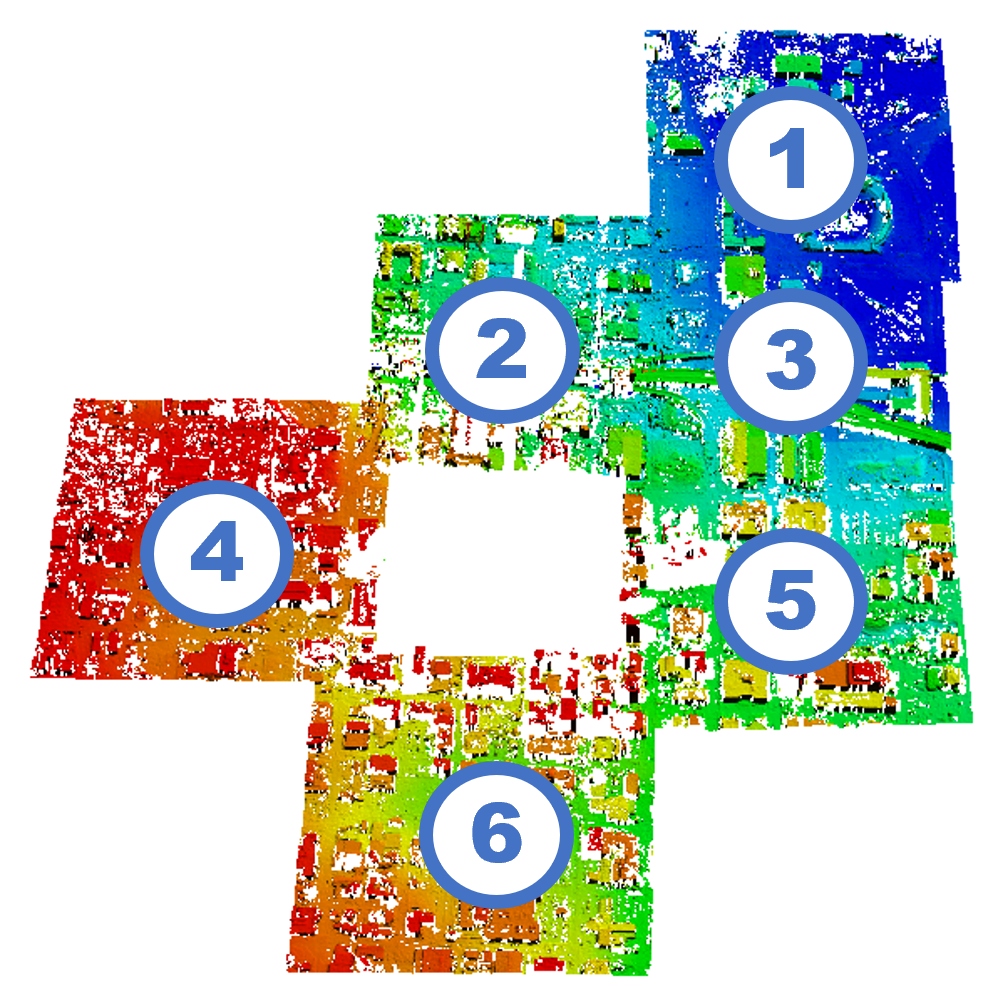}} \hspace{0.05\columnwidth}
        \subcaptionbox{OMA3 : MST graph}{\includegraphics[width=0.25\columnwidth]
        {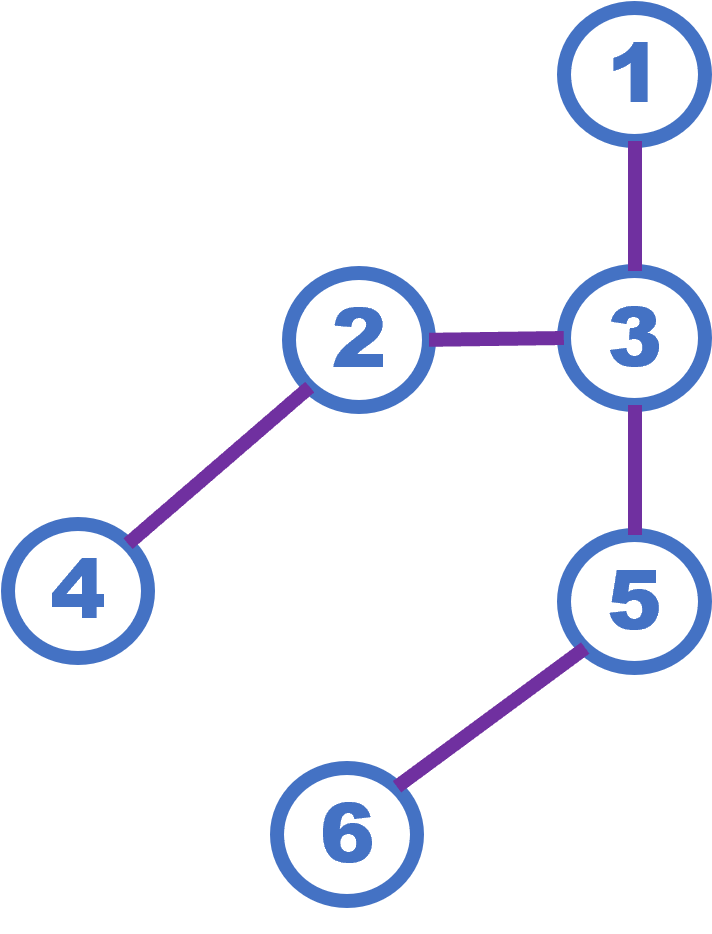}} \hspace{0.05\columnwidth}
        \subcaptionbox{OMA3 : our graph} {\includegraphics[width=0.25\columnwidth]
        {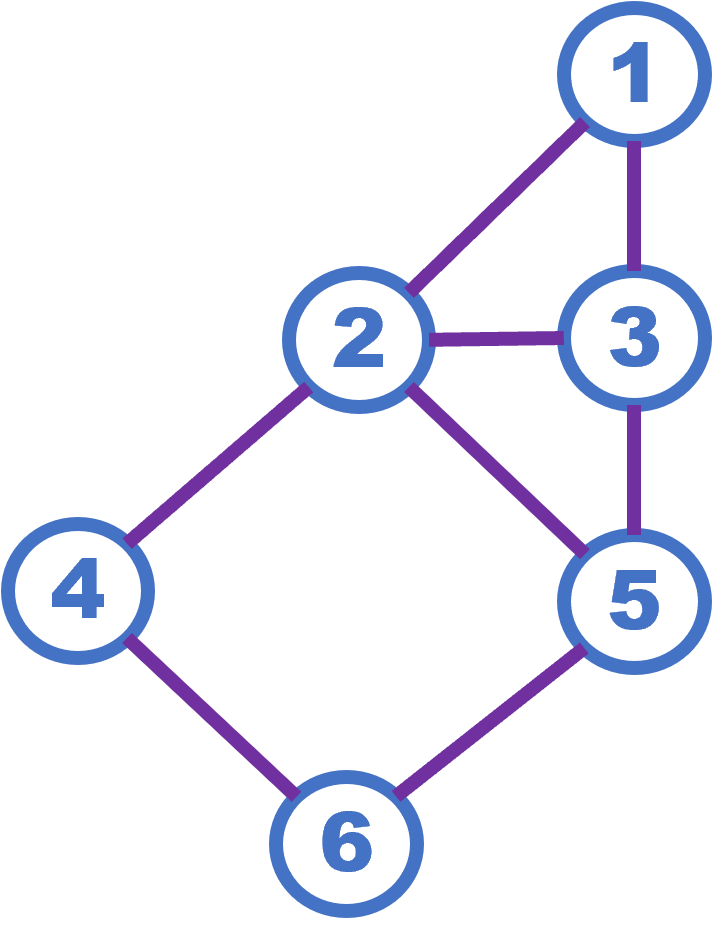}} 
    \caption{Comparison of two approaches for scene graph construction. "MST" denotes the minimal spanning tree, which identifies the minimum number of edges connecting all nodes, while our method establishes edges connecting all possible nodes. In this scenario, the path between any two nodes in our graph is shorter than or equal to those in the MST graph (e.g., node 4\&6 in OMA3), significantly mitigating the accumulated error.}
    \label{fig:graph_construction}
\end{figure}

The primary impediment to the application of k-d tree to large data volumes lies in the necessity to initialize by caching the entire dataset, requiring \(O(NlogN)\) time and \(O(N)\) memory space. A frequently compromised solution involves building a k-d tree for downsampled datasets, albeit at the expense of accuracy reduction. In comparison, our NN search method eliminates the need for initialization. All data are stored on disk, and only local data around the query points are cached. For each NN search, our method demands \(O(k)\) time and memory space, where the upper bound \(d\) is calculated in constant time and \(k\) is the number of pixels within the upper bound. In practice, \(k\ll N\) and as the two DSMs are getting closer, \(k\) will decrease dramatically. Our approach achieves both memory and computational efficiency compared to the k-d tree method. This enables ICP to be applied to large-scale data without compromising accuracy.
\begin{figure*}[tb]
    \centering
        \subcaptionbox{JAX1 (4 DSMs)}{\includegraphics[width=0.32\columnwidth]{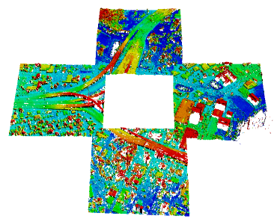}}
        \subcaptionbox{JAX2 (9 DSMs)}{\includegraphics[width=0.32\columnwidth]{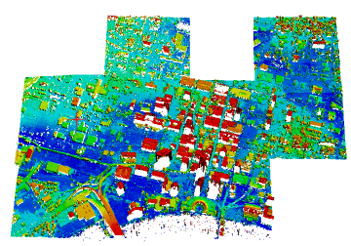}}
        \subcaptionbox{JAX3 (3 DSMs)}{\includegraphics[width=0.32\columnwidth]{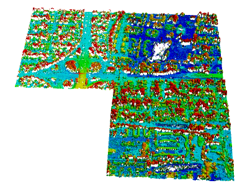}}
        \subcaptionbox{OMA1 (3 DSMs)}{\includegraphics[width=0.32\columnwidth]{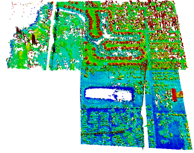}}
        \subcaptionbox{OMA2 (6 DSMs)}{\includegraphics[width=0.32\columnwidth]{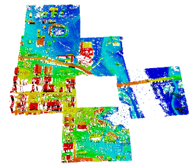}}
        \subcaptionbox{OMA3 (6 DSMs)}{\includegraphics[width=0.32\columnwidth]{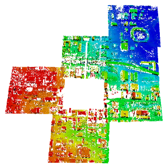}}
        \subcaptionbox{JAX1 lidar}{\includegraphics[width=0.4\columnwidth]{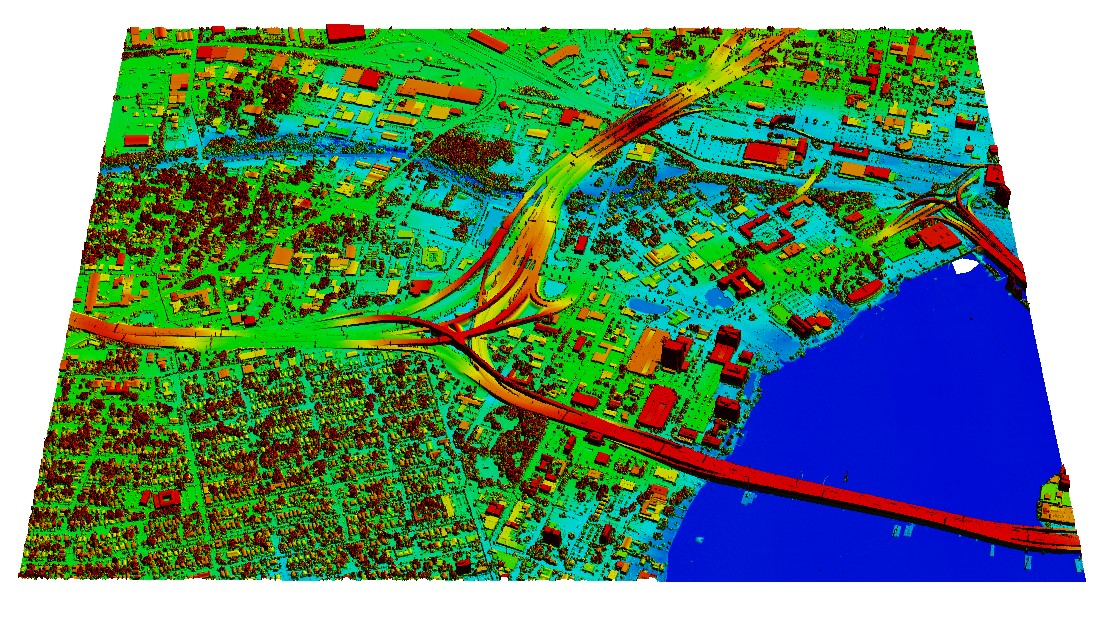}}
        \subcaptionbox{JAX2 lidar}{\includegraphics[width=0.4\columnwidth]{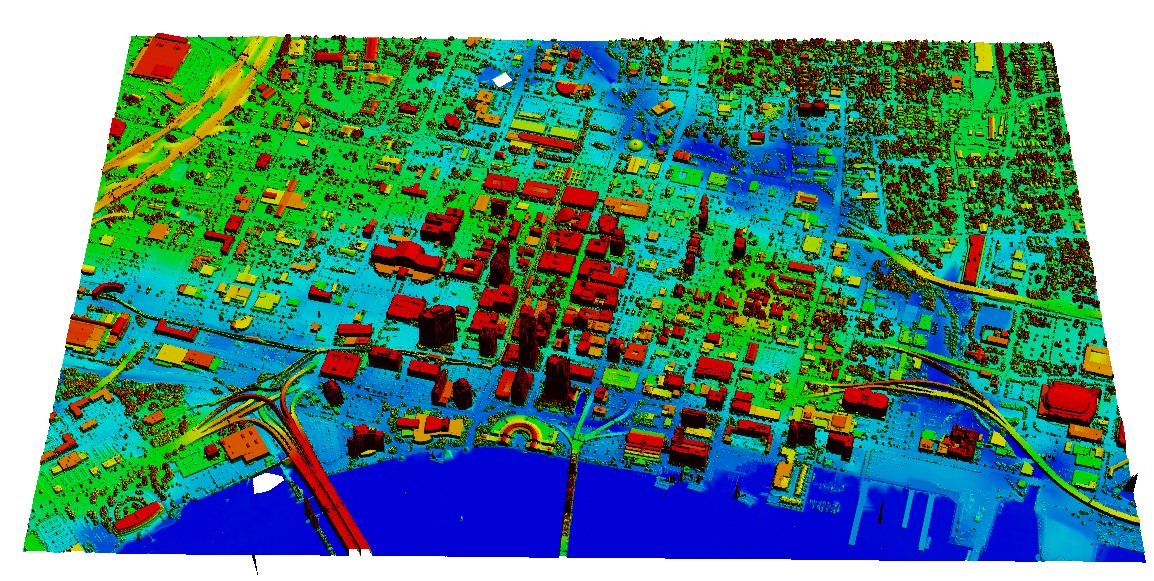}}
        \subcaptionbox{JAX3 LiDAR}{\includegraphics[width=0.4\columnwidth]{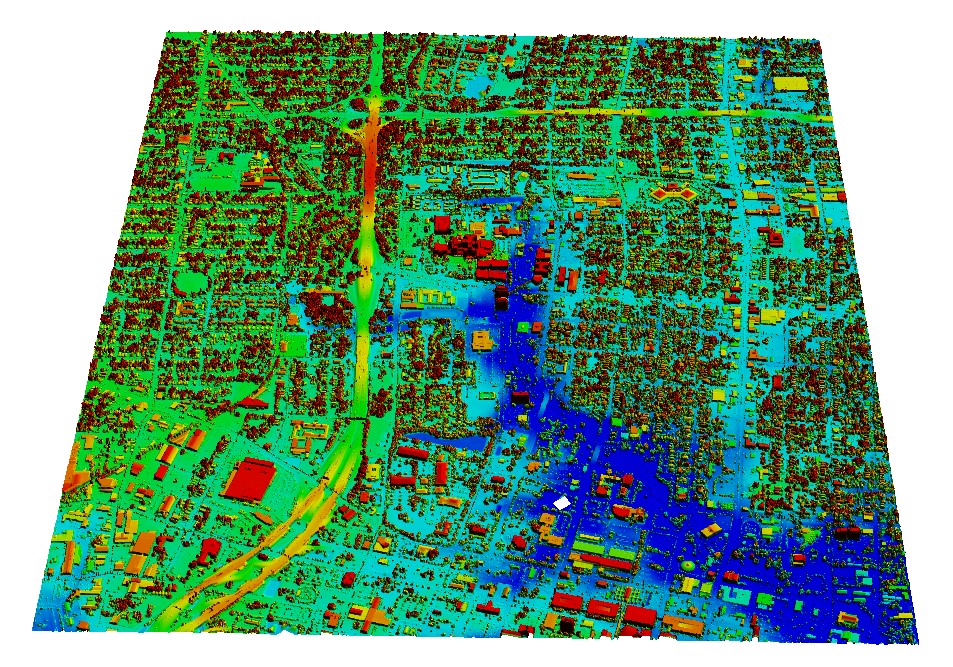}}
        \subcaptionbox{OMA1 lidar}{\includegraphics[width=0.4\columnwidth]{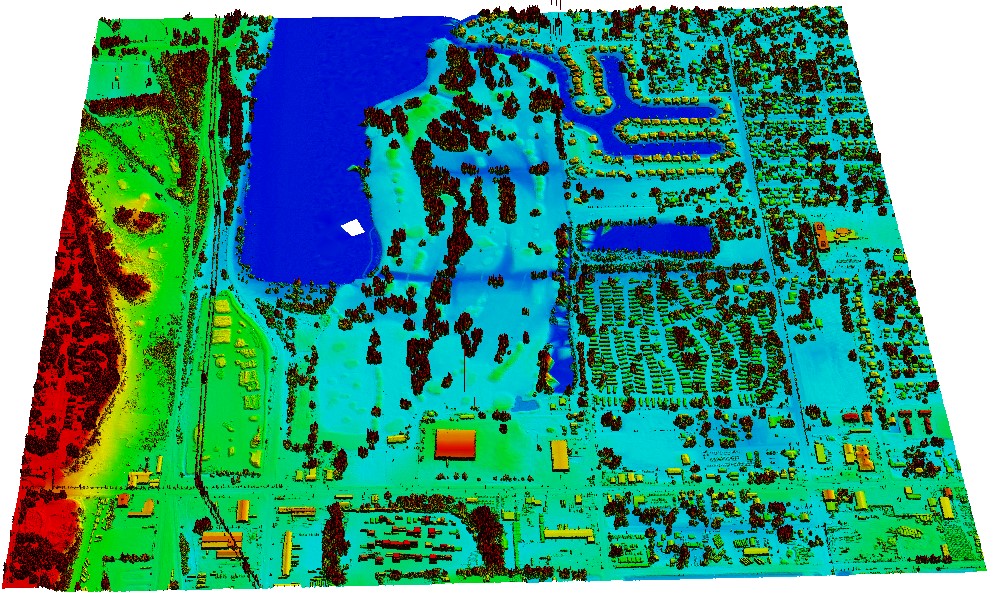}}
        \subcaptionbox{OMA2\&3 lidar}{\includegraphics[width=0.4\columnwidth]{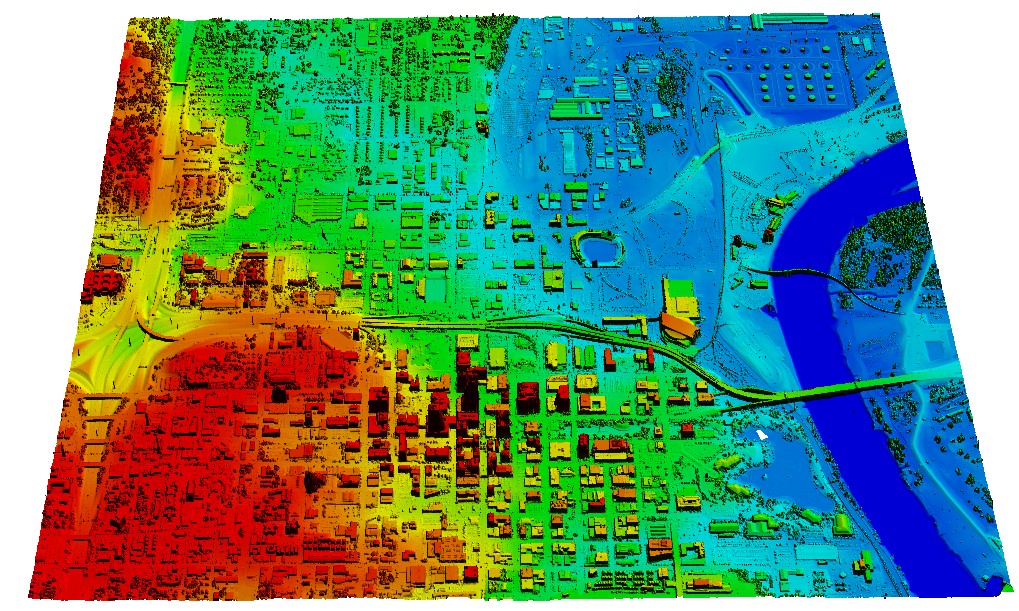}}
    \caption{Illustration of six test sets of satellite DSMs and the ground truth lidar data, where three test sets in Jacksonville area (JAX) and three test sets in Omaha area (OMA). The detailed description is in \autoref{sec:exp_dataset}}
    \label{fig:datasets}
\end{figure*}
\subsection{Motion Averaging-based Framework}\label{sec:32}
\textbf{Scene graph construction.} The scene graph serves as a potent tool for representing pair-wise information among multiple objects and finds extensive application in various engineering problems. In this context, our objective is to construct a scene graph \(\mathcal{G}(\mathcal{V},\mathcal{E})\), where each vertex \(v_i \in \mathcal{V}\) corresponds to a DSM \(P_i\) and each edge \( (i,j) \in \mathcal{E}\) encodes the relative pose between DSM \(P_i\) and \(P_j\).

Existing growing-based methods apply a greedy approach to search for the minimal number of edges connecting all vertex. As the number of vertex increases, the accumulated error between vertex that have a large path cannot be reduced, as shown in \autoref{fig:graph_construction}. A reliable scene graph is intended to capture extensive relative information for providing redundant constraints to make sure any two vertex have a small path. Furthermore, the reliability of each edge must be taken into account to filter out potential outliers. Therefore, for each pair \((P_i,P_j)\), we calculate the overlap score by 
\begin{equation}
    s_{ij}=\frac{\#overlap\:pixels}{\#valid\:pixels}
\end{equation}
where \(s_{ij} \in [0,1]\); Assuming the provided DSMs are approximately geo-referenced, the identification of overlap pixels depends on the presence of corresponding reference pixels at the same geographical location.  Once the \(s_{ij}\) is larger than a pre-defined threshold, the edge will be inserted into the scene graph. Their relative pose \(T_{ij}\) will further be estimated using our DSM-ICP outlined in \autoref{sec:31}.

\textbf{Weight initialization.} The edge reliability can be determined by two factors: the overlap ratio and the quality of pair-wise registration. We associate a weight on each edge to indicate the reliability of the estimated relative pose \(T_{ij}\). It is calculated based on both the overlap score \(s_{ij}\) and the quality score of pair-wise registration \(r_{ij}\):
\begin{equation}
\begin{split}
    w_{ij}=s_{ij}*r_{ij}\\
    r_{ij}=\frac{e^{-err_{ij}}}{\sum_{(i,j) \in \mathcal{E}} e^{-err_{ij}}}
    \end{split}
\end{equation}
where \(err_{ij}\) represents the pair-wise registration error representing the estimation error of registering the correspondences at the last iteration. It can be calculated based on \autoref{eq2} with \(T_{ij}\).

\textbf{Motion averaging.} Given the edge weights and the relative poses \(\{w_{ij}, T_{ij}=(R_{ij},t_{ij}) | (i,j) \in \mathcal{E}\}\), we estimate the global poses \(\{T_{i}=(R_i,t_i)\}\). It is also known as the pose graph optimization problem in the robotics community \cite{carlone2015initialization}. Each relative pose (\(R_{ij}|t_{ij}\)) can be represented by global poses (\(R_{i}|t_{i}\)) and (\(R_{j}|t_{j}\))
\begin{equation}
    t_{ij}=R_i^T(t_j-t_i)+t_{ij}^\epsilon, R_{ij}=R^TR_jR_{ij}^\epsilon
\end{equation}
where the \(t_{ij}^\epsilon \in \mathbb{R}^3, R_{ij}^\epsilon \in SO(3) \) denote the measurement noise. Therefore, we estimate the global poses \(\{R_i,t_i\}\) by solving the optimization problem
\begin{equation}
    \minA_{\substack{\{R_i\} \in SO(3) \\ \{t_i\} \in \mathbb{R}^3}} \sum_{(i,j)\in \mathcal{E}} w_{ij}\|R_{ij}-R_i ^TR_j\|_F^2+w_{ij}\|R_it_{ij}+t_i-t_j\|^2
\end{equation}
where \(\|\cdot\|_F\) means the Frobenius norm of the matrix. The problem can be addressed by decoupling the optimization of rotation and translation. We first estimate the global rotation \({R_i}\) using the closed-form solutions \cite{arie2012global,gojcic2020learning,wang2023robust}. Once the rotation is determined, the translation \({t_i}\) can be computed using the standard least square method \cite{gojcic2020learning}.  

\begin{table*}[tb]
	\centering
		\begin{tabularx}{2\columnwidth}{*{8}{X}} \hline
            \multirow{2}{*}{Method} & \multirow{2}{*}{Scene Graph} & \multicolumn{6}{c}{\(MEAN_{RMSE_{\tau}}\) [m]} \\
			 &  & JAX1 & JAX2 & JAX3 & OMA1 & OMA2 & OMA3 \\ \hline
            -    & -   &  2.213  & 1.715 & 2.632 & 2.231 & 1.877 & 2.126 \\
            Greedy & MST & \textbf{1.832} & 1.514 & \textbf{1.474} & 1.490 & 1.760 & 1.903 \\
            Ours & Full & 1.834 & \textbf{1.424} & \textbf{1.474} & \textbf{1.486} & \textbf{1.727} & \textbf{1.856} \\
            \hline
		\end{tabularx}
	\caption{Quantitative results of accumulated error reduction performance for two methods. "\(MEAN_{RMSE_{\tau}}\)" represents the average \(RMSE_{\tau}\) for all possible pairs. The first row represents the alignment of the raw dataset without any registration process. When the number of DSMs is small (e.g. JAX1, JAX3, and OMA1), the two methods achieve similar performance. When the number of DSMs is large (e.g. JAX2, OMA2, and OMA3), ours achieve better error reduction performance.}
\label{tab:quantitative}
\end{table*}

\section{Experimental Results}\label{sec:experiment}
In this section, two experiments were performed to evaluate the proposed method in terms of: 1) \textbf{pair-wise registration performance}: the memory and computation performance of the proposed DSM-ICP given varying volumes of datasets, introduced in \autoref{sec:exp_pairwise}, and 2) \textbf{multiple DSM registration performance}: the average relative error among all DSMs and the overall error with respect to the ground truth data, introduced in \autoref{sec:exp_multi}. The data and evaluation metrics used in the experiments are introduced in \autoref{sec:exp_dataset}.

\begin{figure}[tb]
    \centering
        \subcaptionbox{\ 0.5 million points}{\includegraphics[width=0.49\columnwidth]{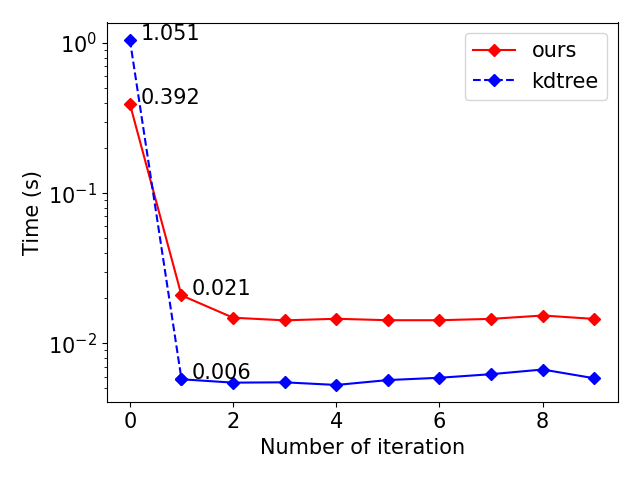}}
        \subcaptionbox{\ 25 million points}{\includegraphics[width=0.49\columnwidth]{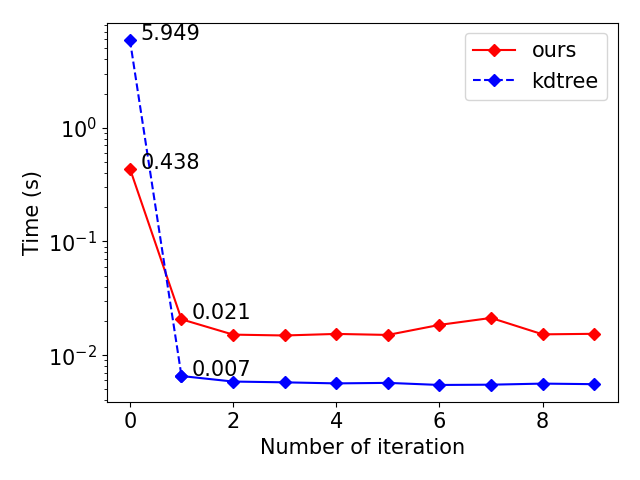}}
        \subcaptionbox{\ 106 million points}{\includegraphics[width=0.49\columnwidth]{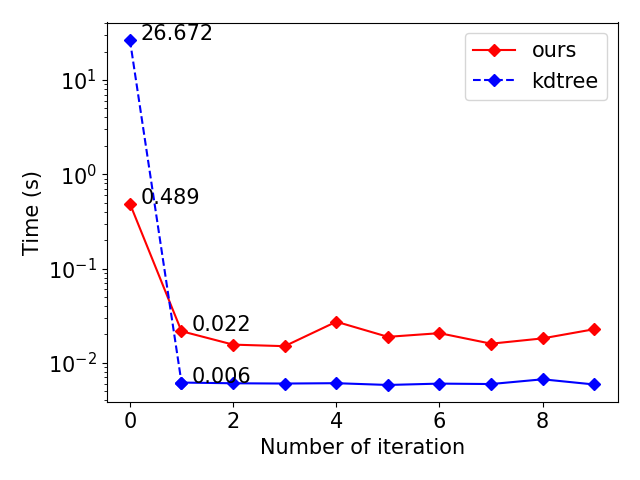}}
        \subcaptionbox{\ 305 million points}{\includegraphics[width=0.49\columnwidth]{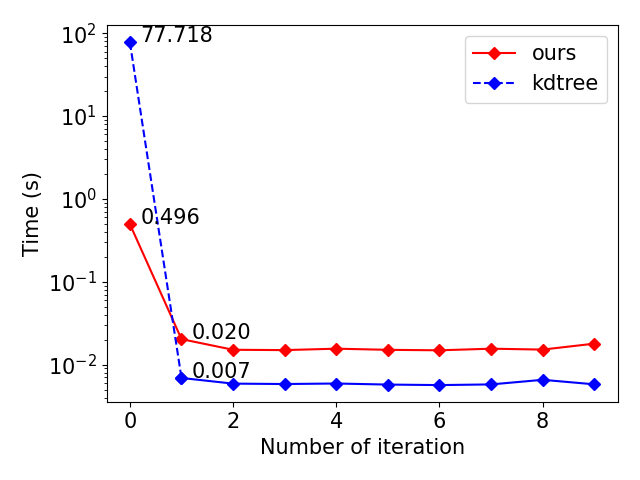}}
    \caption{Computation efficiency of NN search for DSM-ICP and the standard ICP using k-d tree. The y-axis represents the time cost of the NN search at each iteration. In this experiment, 2065 points in total were performed NN search within the varying number of reference points from 0.5 million (a) to 305 million points (d). For ICP using k-d tree, the "Blue dashed" line represents the initialization of k-d tree, followed by NN search.}
    \label{fig:time_eff}
\end{figure}

\begin{figure}[tb]
    \centering
     {\includegraphics[width=\columnwidth]{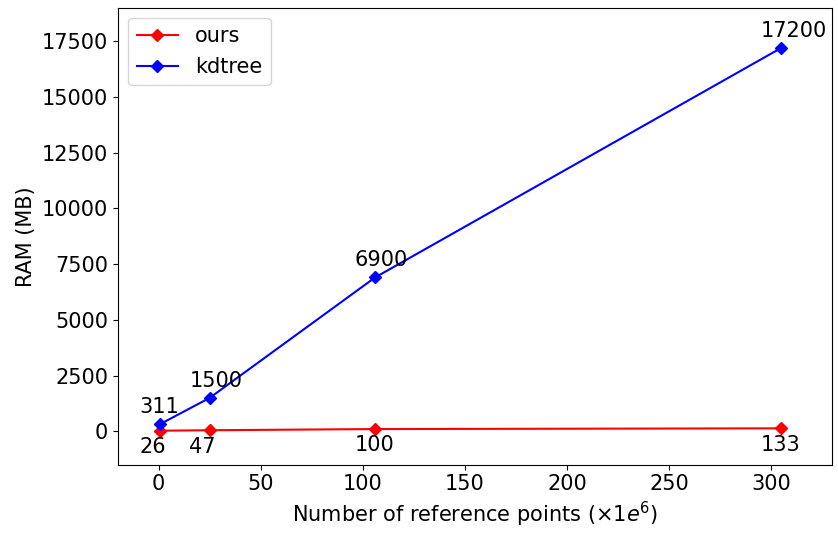}}
    \caption{Memory consumption of the proposed DSM-ICP and the standard ICP using k-d tree with the varying number of reference points. The experiment setting is the same as \autoref{fig:time_eff}}
    \label{fig:memory}
\end{figure}
\subsection{Datasets and Metrics} \label{sec:exp_dataset}
\textbf{Datasets.} We choose the public multi-view satellite datasets: DFC 2019 (tack 3) \cite{c6tm-vw12-19}, encompassing the city of Jacksonville (short for JAX), Florida, USA, and Omaha (short for OMA), Nebraska, USA. These datasets were captured by WorldView-3\footnote{https://resources.maxar.com/data-sheets/worldview-3} with ground-sampling distances of 35cm. In track 3, the satellite images were cropped into sub-images, totaling 118 sub-areas. Each subarea covers \(700m \times 700m\) and is observed by 30 images. The sub-areas partially overlap, and we manually chose six sets of data as the test data, as illustrated in \autoref{fig:datasets}. For each sub-area, we conducted dense matching to generate the DSM using satellite 3D reconstruction software\footnote{https://u.osu.edu/qin.324/rsp/}. The ground truth data for both regions are airborne lidar scanning data from the USGS 3DEP program\footnote{https://www.usgs.gov/3d-elevation-program}, with a point density of 24.64 \(pt/m^2\), 5.82 \(pt/m^2\) in Omaha area. 

\textbf{Metrics.} The pair-wise registration performance can be reflected by the distance between two registered DSMs. In practical scenarios, the standard Root Mean Squared Error (RMSE) may not be effective due to the presence of outlier pixels within DSMs, which can significantly influence overall metrics. Therefore, we implement a straightforward outlier rejection mechanism in conjunction with RMSE:
\begin{equation}
    rmse_{\tau}=\sqrt{\frac{1}{N} \sum_{i\in P} \llbracket \|p_i-q_i\|<\tau \rrbracket (p_i-q_i)^2}
\end{equation}
where the \(\llbracket \cdot \rrbracket\) is the Iverson bracket, \(\tau\) is a pre-defined inlier threshold (10m in our case). \((p_i,q_i)\) is the pair of points corresponding to the same horizontal location. 

\subsection{Performance of Pair-wise Registration} \label{sec:exp_pairwise}
The primary advantage of the proposed DSM-ICP lies in the elimination of the need for initializing spatial data structures such as k-d tree. In each NN search, its cached data is confined to the local area rather than encompassing the entire region. We assess its computational and memory efficiency across diverse volumes of reference data. Specifically, we measure the time cost for every NN search and the overall RAM consumption for both DSM-ICP and the standard ICP utilizing k-d tree.

\begin{table*}[tb]
	\centering
		\begin{tabularx}{2\columnwidth}{*{8}{X}} \hline
            \multirow{2}{*}{Method} & \multirow{2}{*}{Scene Graph} & \multicolumn{6}{c}{\(RMSE_{\tau}\) [m]} \\
             & & JAX1 & JAX2 & JAX3 & OMA1 & OMA2 & OMA3 \\ \hline
			Greedy & MST & 2.305 & 2.166 & 2.756 & 2.065 & 1.461 & 1.667 \\
            Ours & Full & \textbf{2.302} & \textbf{2.129} & 2.756 & 2.065 & \textbf{1.451} & \textbf{1.539}  \\ \hline
		\end{tabularx}
	\caption{Quantitative assessment of the quality of the fused DSM, measured by \(RMSE_{\tau}\) against the lidar ground truth. Our method demonstrates superior reconstruction quality in 4 out of 6 sets. As observed in \autoref{fig:qualit_err}, particularly for scenarios with a large number of DSMs (e.g., JAX2, OMA2, OMA3), our approach outperforms the greedy method significantly.}\label{tab:quant_lidar}
\end{table*}
\begin{figure}[tb]
    \centering
        \subcaptionbox{JAX2 Greedy}{\includegraphics[width=0.4\columnwidth]{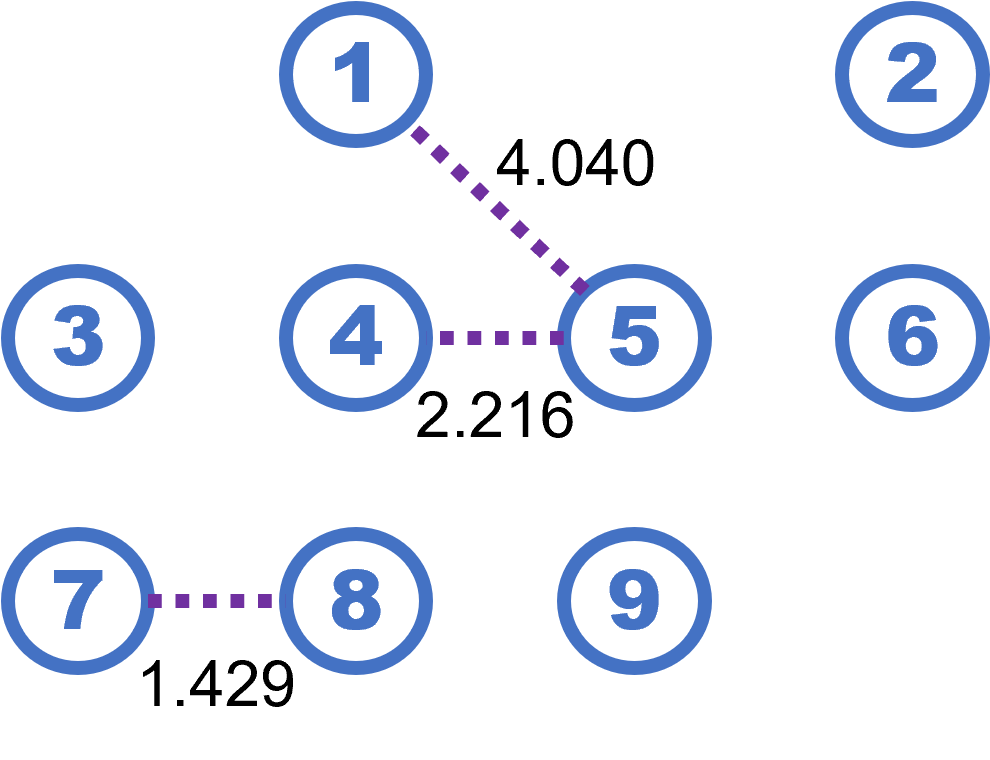}} \hspace{0.1\columnwidth}
        \subcaptionbox{JAX2 Ours}{\includegraphics[width=0.4\columnwidth]{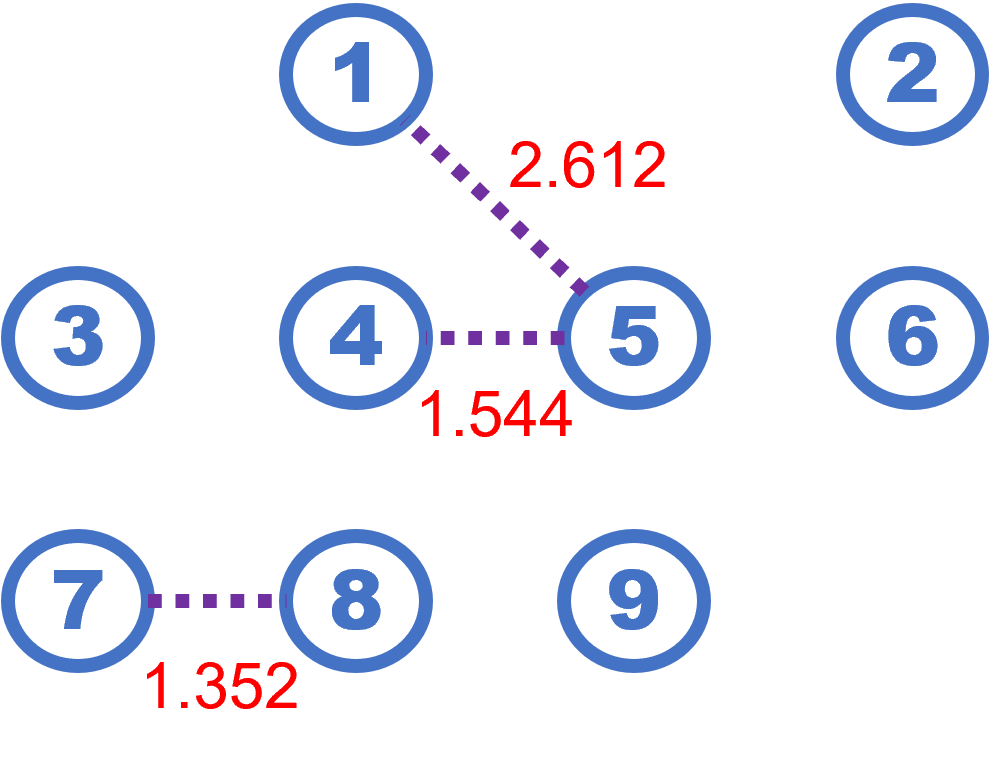}}
        \subcaptionbox{OMA3 Greedy}{\includegraphics[width=0.3\columnwidth]{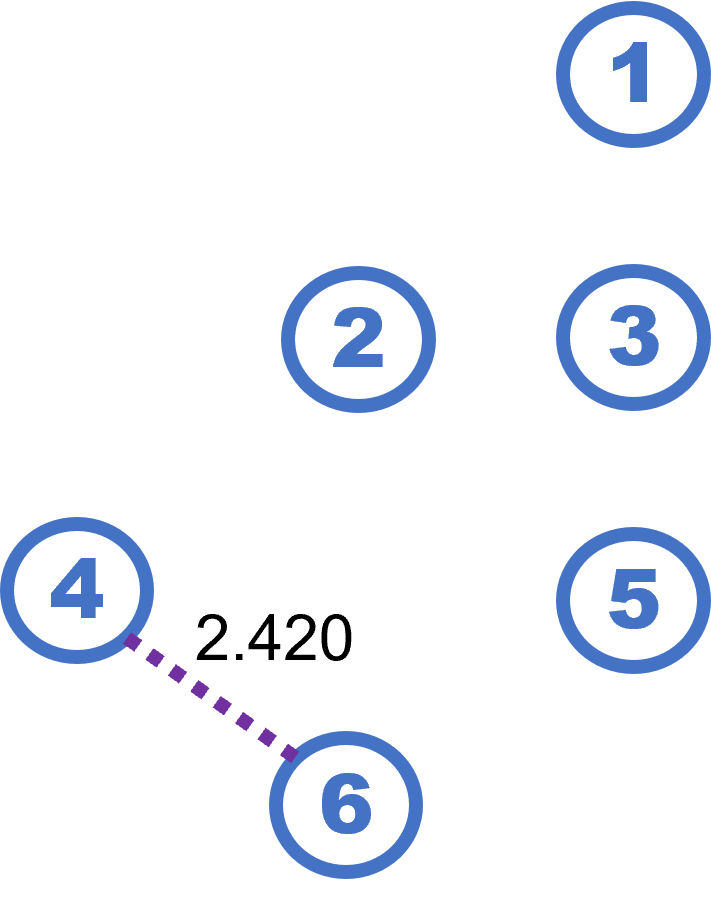}}\hspace{0.2\columnwidth}
        \subcaptionbox{OMA3 Ours}{\includegraphics[width=0.3\columnwidth]{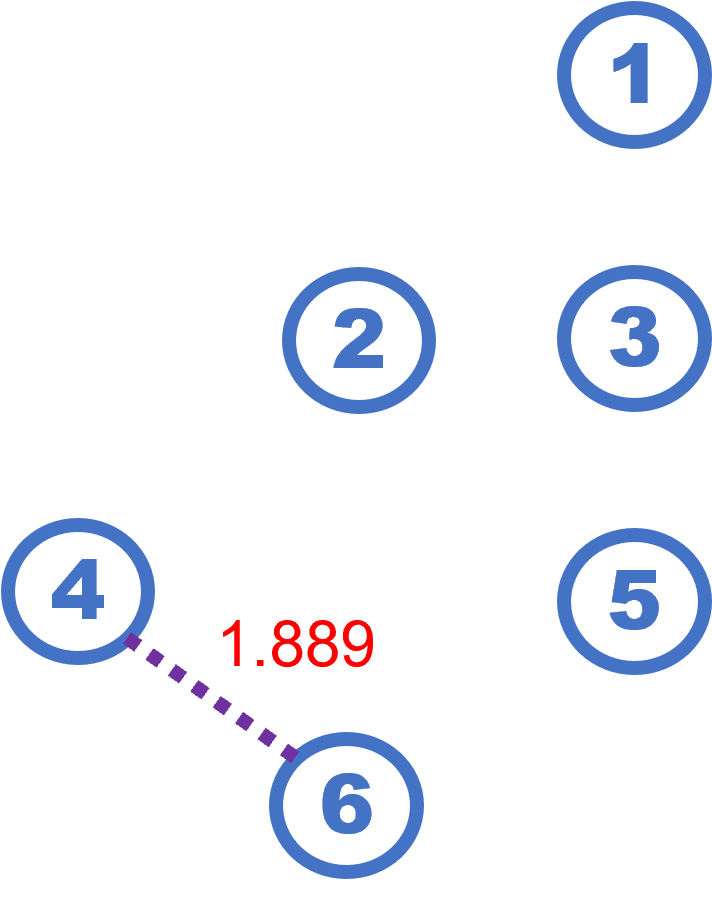}}
    \caption{Registration errors of selected pairs in JAX2 and OMA3 area. The MST and our graphs are shown in \autoref{fig:graph_construction}. We only display the pairs that are likely to accumulate the errors. Two nodes of these pairs are connected by a long path (more than 3 edges) in the MST graph.}
    \label{fig:node_err}
\end{figure}
\begin{figure}[tb]
    \centering
        \subcaptionbox{JAX2 nodes 4\&5 }{\includegraphics[width=0.49\columnwidth]{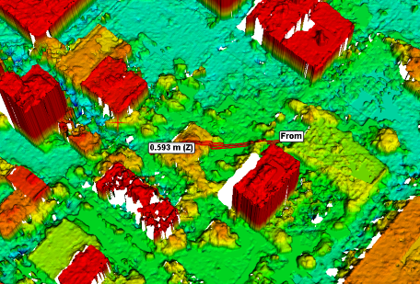}}
        \subcaptionbox{OMA3 nodes 1\&3}{\includegraphics[width=0.49\columnwidth]{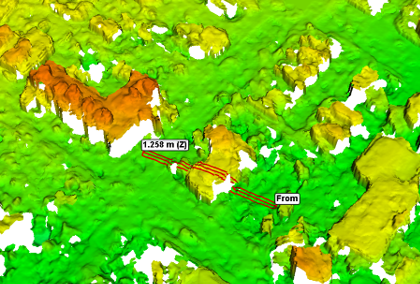}}
        \subcaptionbox{JAX2 profile}{\includegraphics[width=0.49\columnwidth]{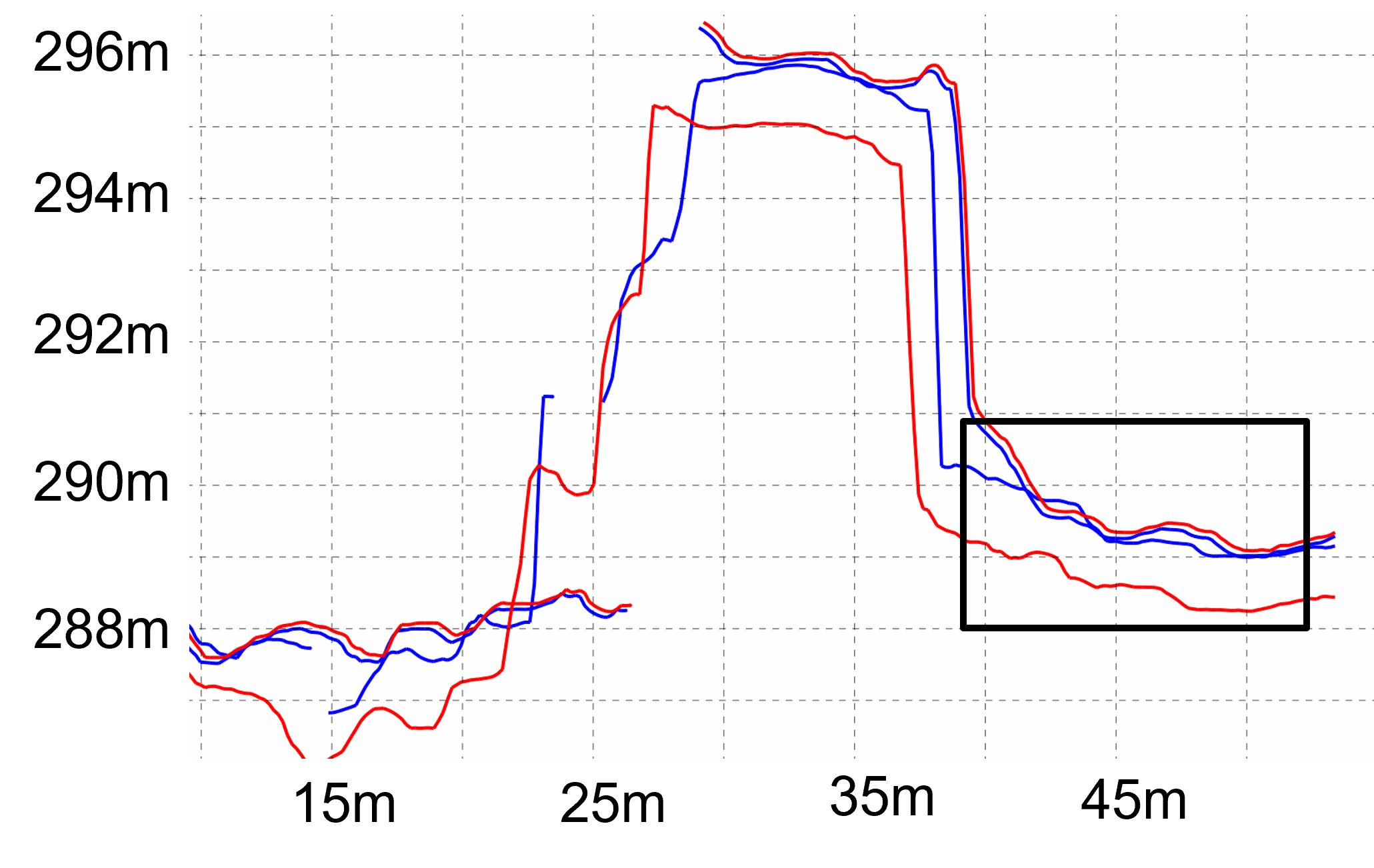}}
        \subcaptionbox{JAX2 zoom in}{\includegraphics[width=0.49\columnwidth]{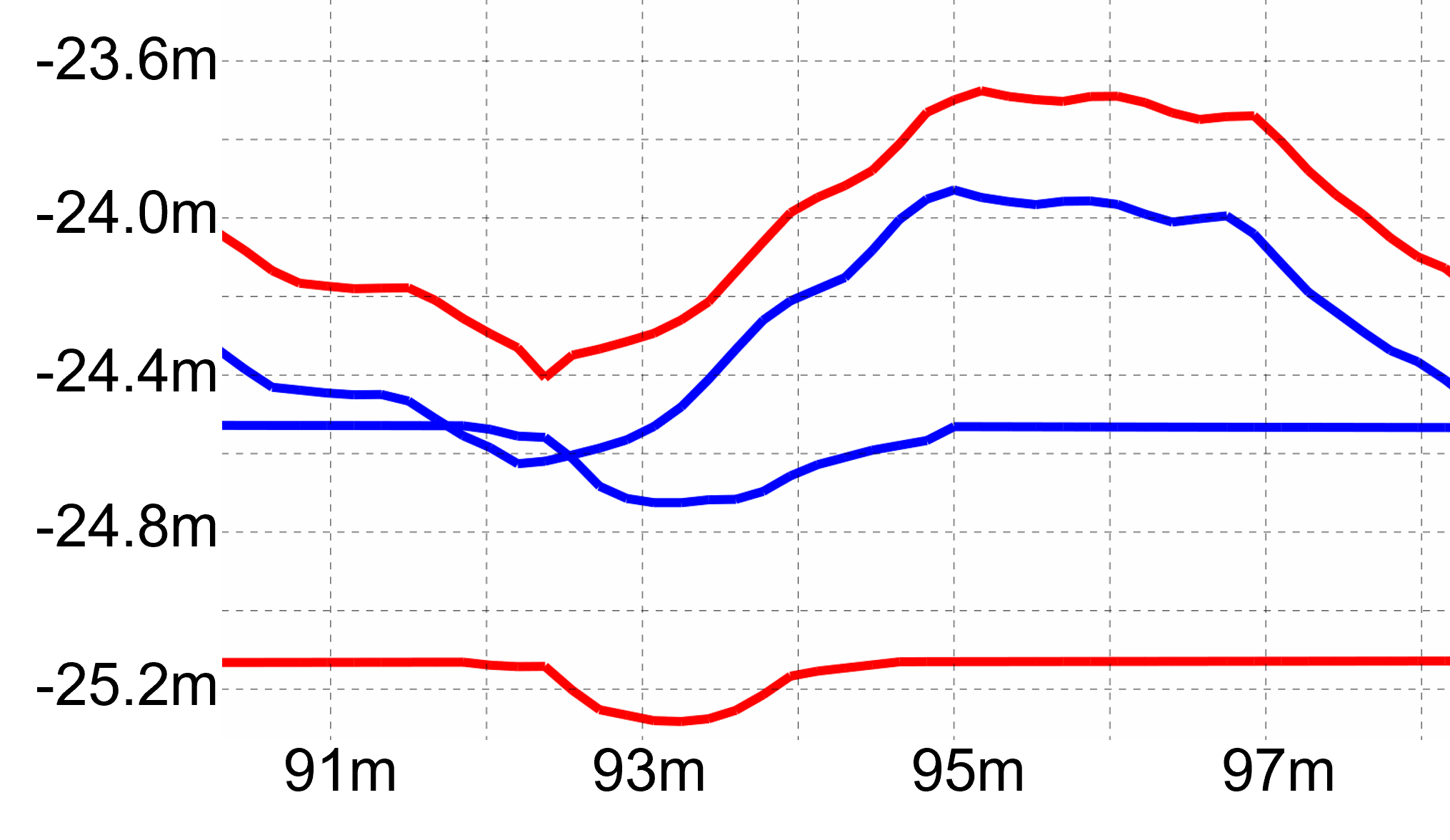}}
        \subcaptionbox{OMA3 profile}{\includegraphics[width=0.49\columnwidth]{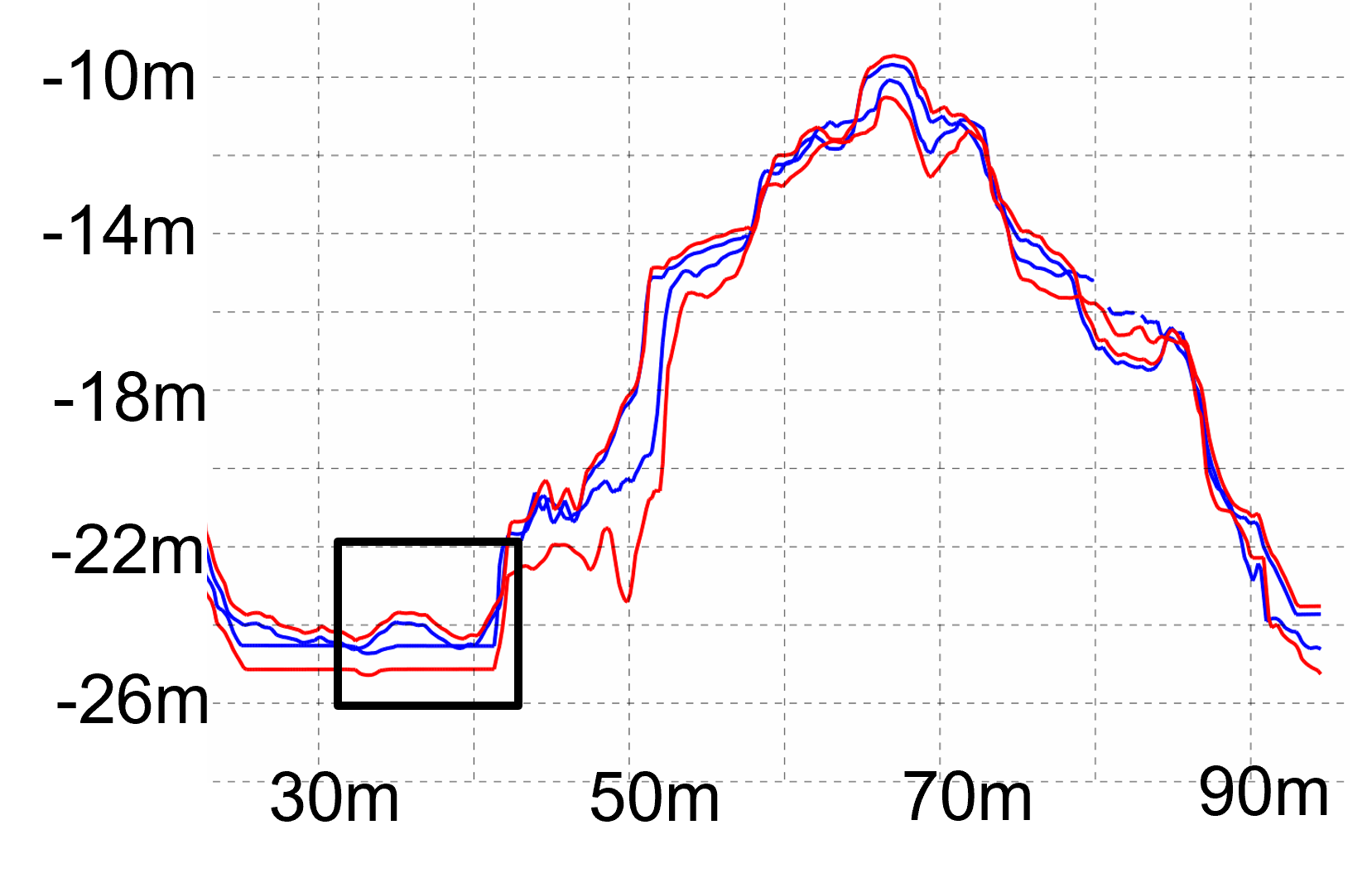}}
        \subcaptionbox{OMA3 zoom in}{\includegraphics[width=0.49\columnwidth]{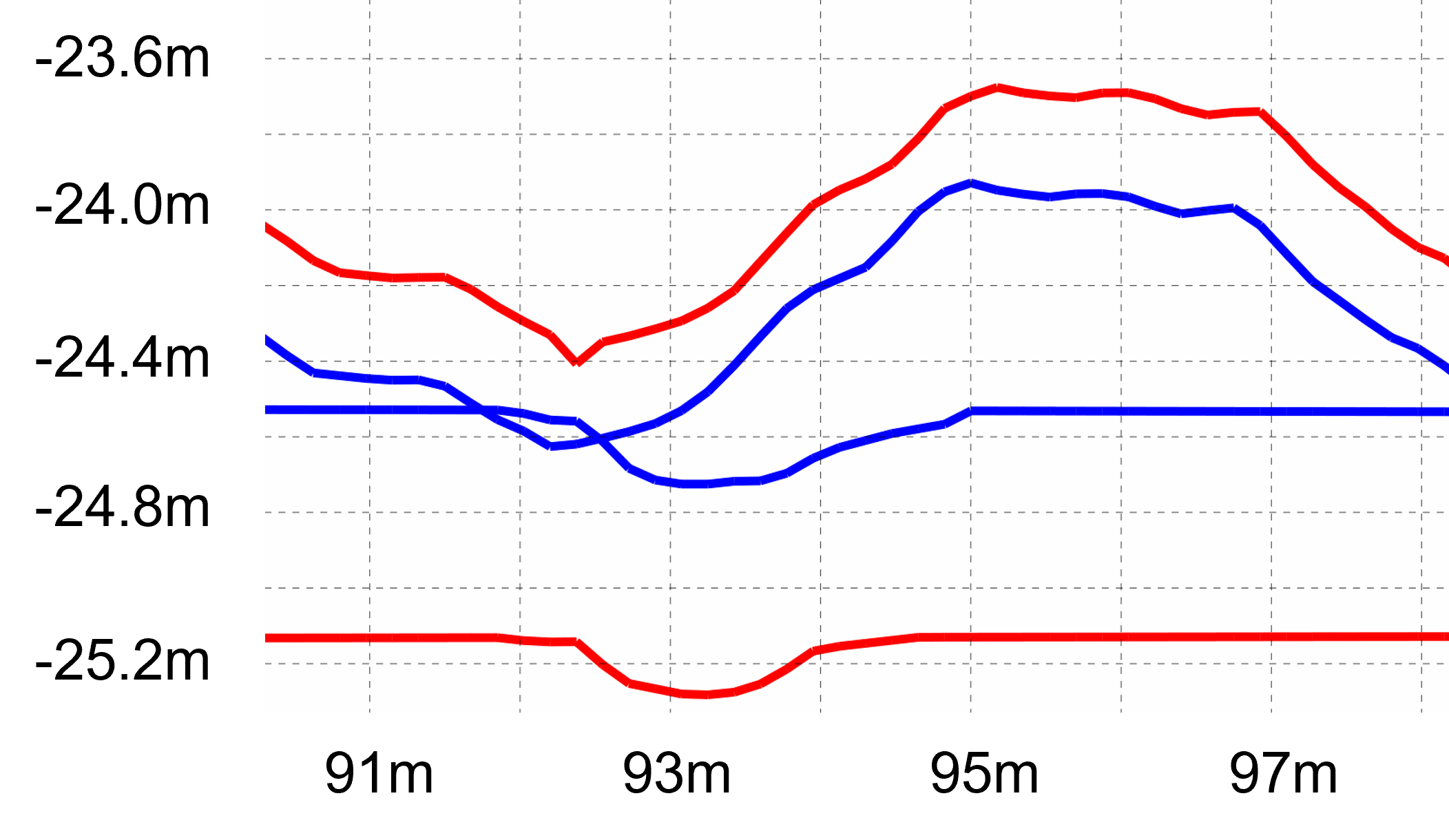}}
    \caption{Profiles of selected buildings in JAX2 and OMA3 area. We show the profiles of registered DSMs by greedy method (red lines) and ours (blue lines).}
    \label{fig:profile}
\end{figure}
\textbf{Computation efficiency performance.} We measure the time cost during the correspondence step (refer to \autoref{sec:31}) at each iteration for both the proposed DSM-ICP and the standard ICP employing k-d tree. Specifically, we fix the number of query points to 2065 while varying the volume of reference data from 0.5 million to 305 million points. The results are illustrated in \autoref{fig:time_eff}. It is evident that ICP using k-d tree initially consumed more time as it traversed all reference points on the local disk to construct the k-d tree in RAM. In contrast, our method bypassed this process. The time cost gradually decreased as the two DSMs moved apart initially, resulting in a larger searching range \(d\) (refer to details in \autoref{sec:31}). As the two DSMs approached each other, the searching range \(d\) decreased, leading to a smaller time cost for the NN search. After several iterations, the time cost for both methods stabilized. Our method required more time due to accessing data via disk, while k-d tree accessed data via RAM.

\textbf{Memory efficiency performance.} We measure the total RAM utilization throughout the entire iteration process with varying volumes of reference data. As depicted in \autoref{fig:memory}, the RAM consumption of the k-d tree exhibits linearity with the number of reference points, as the space complexity of constructing a k-d tree is \(O(N)\), implying it loads all reference points into RAM. In contrast, our RAM consumption remains independent of the volume of reference data. When the number of reference points reached 305 million, our method required only 133 MB, while the k-d tree demanded 17200 MB—129 times larger than ours, revealing the superior memory efficiency of our approach.

\subsection{Performance of Multiple DSM Registration} \label{sec:exp_multi}
We assess the reduction in accumulated error and the reconstruction performance of the proposed graph-based method. The distinctive advantage of our approach lies in the elimination of accumulated registration errors through redundant pair-wise constraints. The conventional approach for registering multiple DSMs employs a greedy method, which initiates registration with the pair exhibiting the largest overlap and subsequently registers the next largest overlap for the remaining DSMs. This can be accomplished using Kruskal's algorithm to find the minimal spanning tree (MST) of an undirected edge-weighted graph. Both the greedy method and our approach utilize the proposed DSM-ICP as the pair-wise registration method.

\textbf{Accumulated error reduction performance.} We apply the proposed method and the greedy method to six test sets (depicted in \autoref{fig:datasets}) and gauge the average \(RMSE_{\tau}\) for all possible pairs as the evaluation metric. As reported in \autoref{tab:quantitative}, our method can reduce initial systematic errors by \(0.2-1.2m\). It achieved the highest registration accuracy in 5 out of 6 sets. The greedy algorithm performs well when the graph size is small; however, as the graph enlarges, relative registration errors will accumulate along the path connecting two distant nodes. The advantage of our method lies in the provision of more pair-wise constraints, offering additional directions to distribute the accumulated error across the graph.

Specifically, we select two test sets and present their pair-wise \(RMSE_{\tau}\) in \autoref{fig:node_err}. Nodes 4 and 6 in OMA3 are four steps away from each other in the MST graph and their \(RMSE_{\tau}\) is 2.420m. In comparison, ours reduced the registration error to 1.889m. The same holds for the nodes 1\&5, 4\&5, and 7\&8 in JAX2. 

Additionally, we showcase the qualitative results of nodes 4 \& 5 of JAX2 and nodes 4 \& 6 of OMA3 by illustrating the profiles of registered DSMs in \autoref{fig:profile}. In the JAX2 area, DSMs registered by the greedy method exhibit a rough 1m height difference, whereas ours nearly perfectly aligns the two DSMs, considering the quality of the DSM. Similarly, our method achieves nearly perfect registration of DSMs in OMA3, while the greedy method results in a rough 1m height difference.

\begin{figure}
    \centering
        \subcaptionbox{JAX2 Greedy }{\includegraphics[width=0.49\columnwidth]{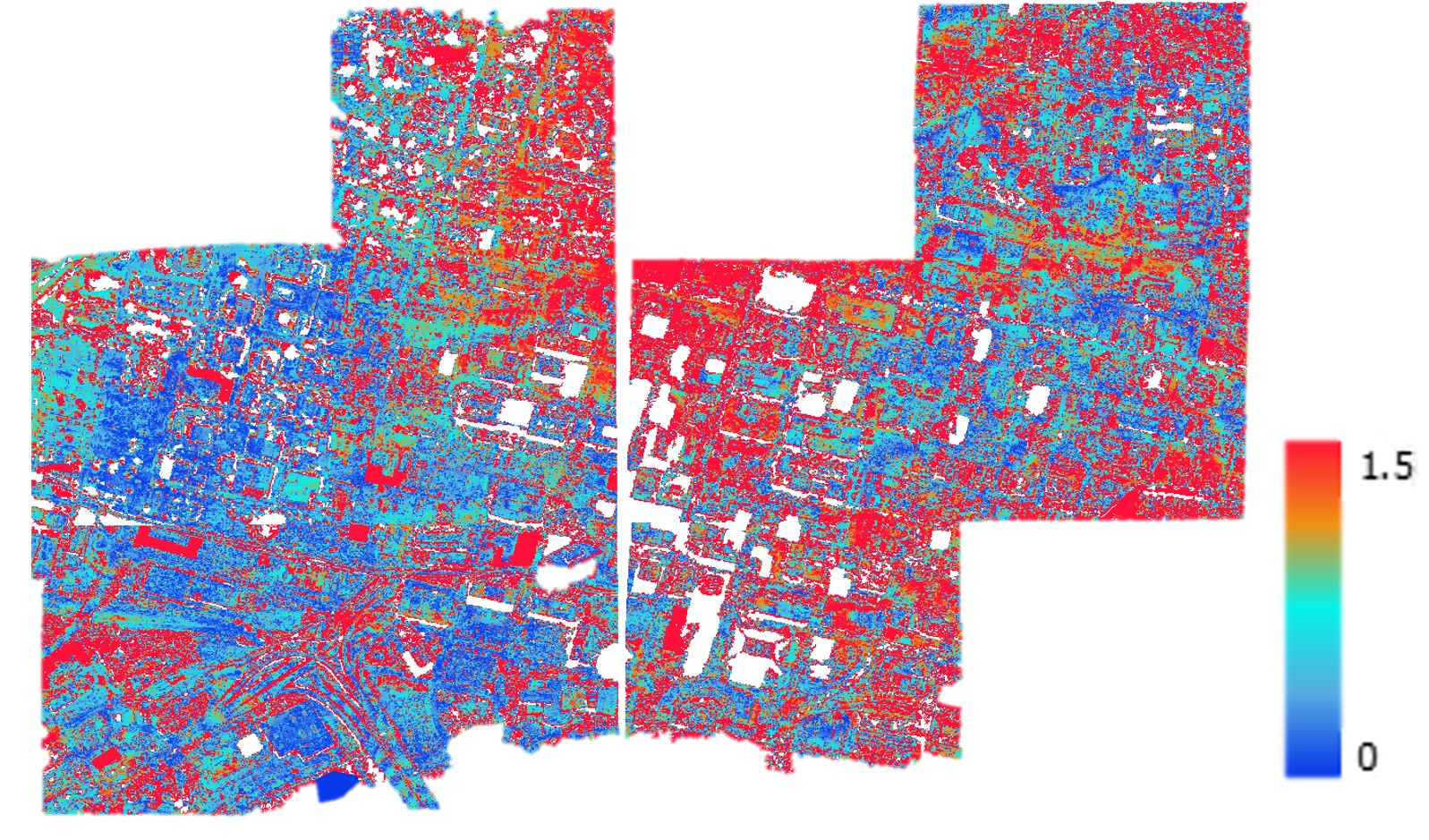}}
        \subcaptionbox{JAX2 Ours}{\includegraphics[width=0.49\columnwidth]{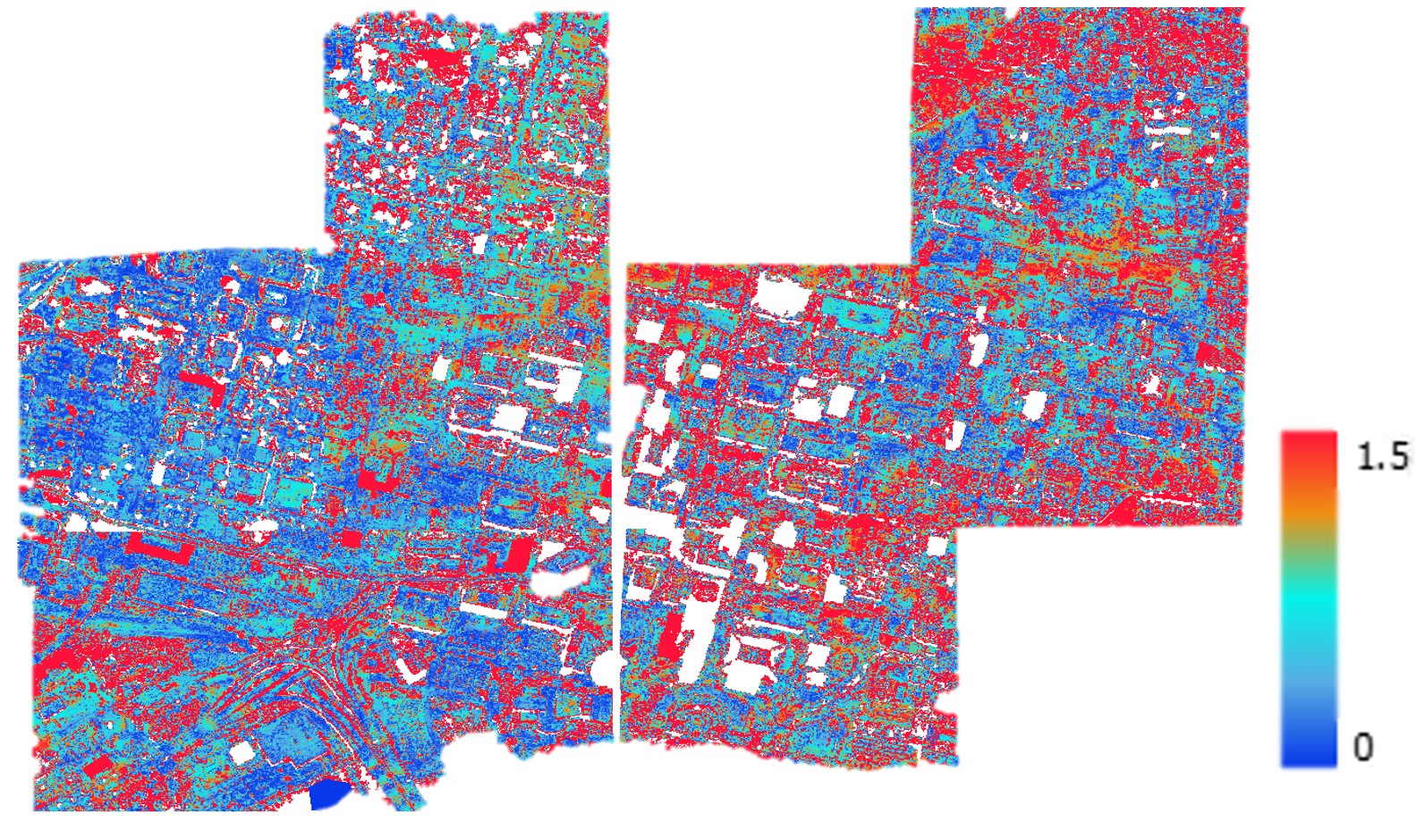}}
        \subcaptionbox{OMA3 Greedy}{\includegraphics[width=0.49\columnwidth]{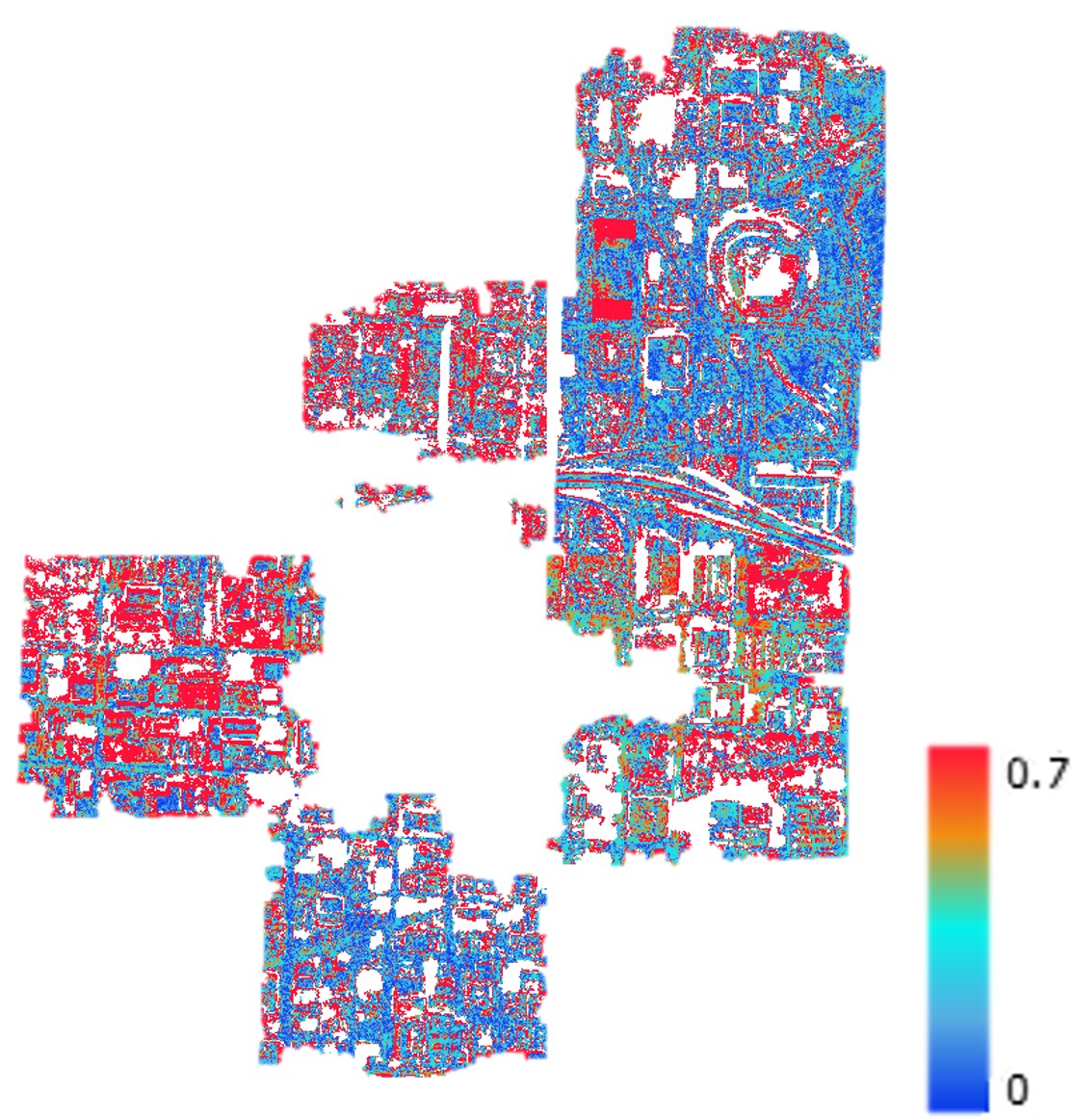}}
        \subcaptionbox{OMA3 Ours}{\includegraphics[width=0.49\columnwidth]{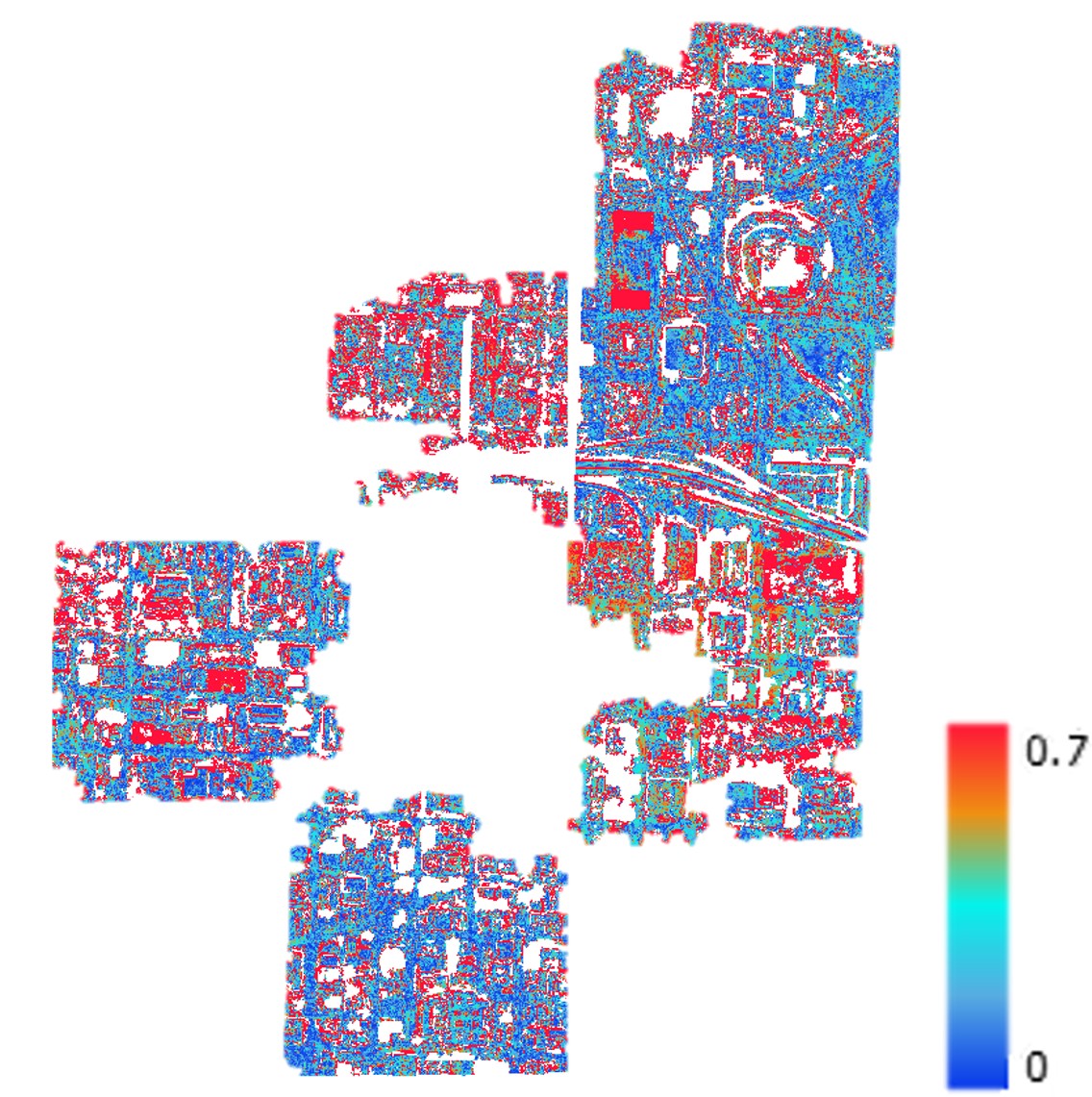}}
    \caption{Qualitative results of error map depicting the height difference between the fused DSM and the lidar ground truth. The greedy method produces more error data than ours, particularly around the middle-top part in JAX2 and the left part in OMA3.}
    \label{fig:qualit_err}
\end{figure}

\textbf{Reconstruction performance.} After eliminating systematic errors among the DSMs, we fuse all registered DSMs into a single DSM and assess its quality against lidar ground truth. Specifically, we align the fused DSM (achieved using RSP) with the ground truth using DSM-ICP and compute \(RMSE_{\tau}\). The quantitative results are presented in \autoref{tab:quant_lidar}, wherein our approach achieved superior reconstruction accuracy in 4 out of 6 areas. Moreover, we showcase qualitative results illustrating the quality of the fused DSM in two selected areas, JAX2 and OMA3, as depicted in \autoref{fig:qualit_err}. Regions with significant errors are colorized in "Red". In the JAX2 area, large errors produced by the greedy method are mainly concentrated around the middle part, where accumulated errors between nodes 4\&5 and 1\&5 (see \autoref{fig:node_err}) were not effectively reduced by the greedy method. Similar observations apply to the left part of the OMA3 area.

\section{Conclusion}\label{sec:conclusion}
In this paper, we presented a motion averaging-based approach for the registration of multiple large-scale DSMs. Specifically, we proposed a fast and exact NN search method by utilizing the grid structure of DSM. Its computation and memory complexity are independent of the volume of the input data, enabling our pair-wise registration method, DSM-ICP, to be scalable to large-scale data. Based on it, we perform pair-wise registration for all possible pairs of DSMs to construct a reliable scene graph, where the edge is weighted based on overlap and registration error. Then, the global poses for each DSM are estimated by redistributing the pair-wise registration errors over the whole scene graph. We evaluated the proposed method on public satellite datasets with the ground truth airborne lidar data. The experiment results demonstrate that our DSM-ICP achieves both superior computation and memory efficiency than k-d tree-based ICP. The proposed motion averaging-based method achieves better error reduction performance and reconstruction performance than the greedy method.
{
    \small
    \bibliographystyle{ieeenat_fullname}
    \bibliography{main}

\begin{thebibliography}{30}
\providecommand{\natexlab}[1]{#1}
\providecommand{\url}[1]{\texttt{#1}}
\expandafter\ifx\csname urlstyle\endcsname\relax
  \providecommand{\doi}[1]{doi: #1}\else
  \providecommand{\doi}{doi: \begingroup \urlstyle{rm}\Url}\fi

\bibitem[Ao et~al.(2021)Ao, Hu, Yang, Markham, and Guo]{ao2021spinnet}
Sheng Ao, Qingyong Hu, Bo Yang, Andrew Markham, and Yulan Guo.
\newblock Spinnet: Learning a general surface descriptor for 3d point cloud registration.
\newblock In \emph{Proceedings of the IEEE/CVF conference on computer vision and pattern recognition}, pages 11753--11762, 2021.

\bibitem[Arie-Nachimson et~al.(2012)Arie-Nachimson, Kovalsky, Kemelmacher-Shlizerman, Singer, and Basri]{arie2012global}
Mica Arie-Nachimson, Shahar~Z Kovalsky, Ira Kemelmacher-Shlizerman, Amit Singer, and Ronen Basri.
\newblock Global motion estimation from point matches.
\newblock In \emph{2012 Second international conference on 3D imaging, modeling, processing, visualization \& transmission}, pages 81--88. IEEE, 2012.

\bibitem[Arun et~al.(1987)Arun, Huang, and Blostein]{arun1987least}
K~Somani Arun, Thomas~S Huang, and Steven~D Blostein.
\newblock Least-squares fitting of two 3-d point sets.
\newblock \emph{IEEE Transactions on pattern analysis and machine intelligence}, \penalty0 (5):\penalty0 698--700, 1987.

\bibitem[Bai et~al.(2021)Bai, Luo, Zhou, Chen, Li, Hu, Fu, and Tai]{bai2021pointdsc}
Xuyang Bai, Zixin Luo, Lei Zhou, Hongkai Chen, Lei Li, Zeyu Hu, Hongbo Fu, and Chiew-Lan Tai.
\newblock Pointdsc: Robust point cloud registration using deep spatial consistency.
\newblock In \emph{Proceedings of the IEEE/CVF Conference on Computer Vision and Pattern Recognition}, pages 15859--15869, 2021.

\bibitem[Barath and Matas(2018)]{barath2018graph}
Daniel Barath and Ji{\v{r}}{\'\i} Matas.
\newblock Graph-cut ransac.
\newblock In \emph{Proceedings of the IEEE conference on computer vision and pattern recognition}, pages 6733--6741, 2018.

\bibitem[Besl and McKay(1992)]{besl1992method}
Paul~J Besl and Neil~D McKay.
\newblock Method for registration of 3-d shapes.
\newblock In \emph{Sensor fusion IV: control paradigms and data structures}, pages 586--606. Spie, 1992.

\bibitem[Bouaziz et~al.(2013)Bouaziz, Tagliasacchi, and Pauly]{bouaziz2013sparse}
Sofien Bouaziz, Andrea Tagliasacchi, and Mark Pauly.
\newblock Sparse iterative closest point.
\newblock In \emph{Computer graphics forum}, pages 113--123. Wiley Online Library, 2013.

\bibitem[Carlone et~al.(2015)Carlone, Tron, Daniilidis, and Dellaert]{carlone2015initialization}
Luca Carlone, Roberto Tron, Kostas Daniilidis, and Frank Dellaert.
\newblock Initialization techniques for 3d slam: A survey on rotation estimation and its use in pose graph optimization.
\newblock In \emph{2015 IEEE international conference on robotics and automation (ICRA)}, pages 4597--4604. IEEE, 2015.

\bibitem[Chen et~al.(2019)Chen, Nan, Xia, Zhao, and Wonka]{chen2019plade}
Songlin Chen, Liangliang Nan, Renbo Xia, Jibin Zhao, and Peter Wonka.
\newblock Plade: A plane-based descriptor for point cloud registration with small overlap.
\newblock \emph{IEEE Transactions on Geoscience and Remote Sensing}, 58\penalty0 (4):\penalty0 2530--2540, 2019.

\bibitem[Chen and Yu(2019)]{chen2019feature}
Xijiang Chen and Kegen Yu.
\newblock Feature line generation and regularization from point clouds.
\newblock \emph{IEEE Transactions on Geoscience and Remote Sensing}, 57\penalty0 (12):\penalty0 9779--9790, 2019.

\bibitem[Eggert and Dalyot(2012)]{eggert2012octree}
D Eggert and S Dalyot.
\newblock Octree-based simd strategy for icp registration and alignment of 3d point clouds.
\newblock \emph{ISPRS Annals of the Photogrammetry, Remote Sensing and Spatial Information Sciences; I-3}, 1\penalty0 (1):\penalty0 105--110, 2012.

\bibitem[Gojcic et~al.(2020)Gojcic, Zhou, Wegner, Guibas, and Birdal]{gojcic2020learning}
Zan Gojcic, Caifa Zhou, Jan~D Wegner, Leonidas~J Guibas, and Tolga Birdal.
\newblock Learning multiview 3d point cloud registration.
\newblock In \emph{Proceedings of the IEEE/CVF conference on computer vision and pattern recognition}, pages 1759--1769, 2020.

\bibitem[Greenspan and Yurick(2003)]{greenspan2003approximate}
Michael Greenspan and Mike Yurick.
\newblock Approximate kd tree search for efficient icp.
\newblock In \emph{Fourth International Conference on 3-D Digital Imaging and Modeling, 2003. 3DIM 2003. Proceedings.}, pages 442--448. IEEE, 2003.

\bibitem[Huang and Qin(2023)]{huang2023critical}
Debao Huang and Rongjun Qin.
\newblock a critical analysis of internal reliability for uncertainty quantification of dense image matching in multi-view stereo.
\newblock \emph{arXiv preprint arXiv:2309.09379}, 2023.

\bibitem[Huang et~al.(2022)Huang, Tang, and Qin]{huang2022evaluation}
Debao Huang, Yang Tang, and Rongjun Qin.
\newblock An evaluation of planetscope images for 3d reconstruction and change detection--experimental validations with case studies.
\newblock \emph{GIScience \& Remote Sensing}, 59\penalty0 (1):\penalty0 744--761, 2022.

\bibitem[JPL(2013)]{srtm}
NASA JPL.
\newblock Nasa shuttle radar topography mission global 1 arc second [data set], 2013.

\bibitem[Kruskal(1956)]{kruskal1956shortest}
Joseph~B Kruskal.
\newblock On the shortest spanning subtree of a graph and the traveling salesman problem.
\newblock \emph{Proceedings of the American Mathematical society}, 7\penalty0 (1):\penalty0 48--50, 1956.

\bibitem[Le~Saux et~al.(2019)Le~Saux, Yokoya, Hänsch, and Brown]{c6tm-vw12-19}
Bertrand Le~Saux, Naoto Yokoya, Ronny Hänsch, and Myron Brown.
\newblock Data fusion contest 2019 (dfc2019), 2019.

\bibitem[Mellado et~al.(2014)Mellado, Aiger, and Mitra]{mellado2014super}
Nicolas Mellado, Dror Aiger, and Niloy~J Mitra.
\newblock Super 4pcs fast global pointcloud registration via smart indexing.
\newblock In \emph{Computer graphics forum}, pages 205--215. Wiley Online Library, 2014.

\bibitem[Pavlov et~al.(2018)Pavlov, Ovchinnikov, Derbyshev, Tsetserukou, and Oseledets]{pavlov2018aa}
Artem~L Pavlov, Grigory~WV Ovchinnikov, Dmitry~Yu Derbyshev, Dzmitry Tsetserukou, and Ivan~V Oseledets.
\newblock Aa-icp: Iterative closest point with anderson acceleration.
\newblock In \emph{2018 IEEE International Conference on Robotics and Automation (ICRA)}, pages 3407--3412. IEEE, 2018.

\bibitem[Qin and Gui(2023)]{qin2023using}
Rongjun Qin and Shengxi Gui.
\newblock Using planetscope-derived time-series elevation models to track surging glacier 3d dynamics in mid-latitude mountain regions.
\newblock \emph{AGU23}, 2023.

\bibitem[Rusinkiewicz(2019)]{rusinkiewicz2019symmetric}
Szymon Rusinkiewicz.
\newblock A symmetric objective function for icp.
\newblock \emph{ACM Transactions on Graphics (TOG)}, 38\penalty0 (4):\penalty0 1--7, 2019.

\bibitem[Rusu et~al.(2009)Rusu, Blodow, and Beetz]{rusu2009fast}
Radu~Bogdan Rusu, Nico Blodow, and Michael Beetz.
\newblock Fast point feature histograms (fpfh) for 3d registration.
\newblock In \emph{2009 IEEE international conference on robotics and automation}, pages 3212--3217. IEEE, 2009.

\bibitem[Wang et~al.(2023)Wang, Liu, Dong, Guo, Liu, Wang, and Yang]{wang2023robust}
Haiping Wang, Yuan Liu, Zhen Dong, Yulan Guo, Yu-Shen Liu, Wenping Wang, and Bisheng Yang.
\newblock Robust multiview point cloud registration with reliable pose graph initialization and history reweighting.
\newblock In \emph{Proceedings of the IEEE/CVF Conference on Computer Vision and Pattern Recognition}, pages 9506--9515, 2023.

\bibitem[Xu et~al.(2021)Xu, Huang, Song, Ling, Strasbaugh, Yilmaz, Sezen, and Qin]{xu2021volumetric}
Ningli Xu, Debao Huang, Shuang Song, Xiao Ling, Chris Strasbaugh, Alper Yilmaz, Halil Sezen, and Rongjun Qin.
\newblock A volumetric change detection framework using uav oblique photogrammetry--a case study of ultra-high-resolution monitoring of progressive building collapse.
\newblock \emph{International Journal of Digital Earth}, 14\penalty0 (11):\penalty0 1705--1720, 2021.

\bibitem[Xu et~al.(2023)Xu, Qin, and Song]{xu2023point}
Ningli Xu, Rongjun Qin, and Shuang Song.
\newblock Point cloud registration for lidar and photogrammetric data: A critical synthesis and performance analysis on classic and deep learning algorithms.
\newblock \emph{ISPRS open journal of photogrammetry and remote sensing}, page 100032, 2023.

\bibitem[Xu et~al.(2024)Xu, Qin, Huang, and Remondino]{xu2024multi}
Ningli Xu, Rongjun Qin, Debao Huang, and Fabio Remondino.
\newblock Multi-tiling neural radiance field (nerf)—geometric assessment on large-scale aerial datasets.
\newblock \emph{The Photogrammetric Record}, 2024.

\bibitem[Yang et~al.(2020)Yang, Shi, and Carlone]{yang2020teaser}
Heng Yang, Jingnan Shi, and Luca Carlone.
\newblock Teaser: Fast and certifiable point cloud registration.
\newblock \emph{IEEE Transactions on Robotics}, 37\penalty0 (2):\penalty0 314--333, 2020.

\bibitem[Zhang et~al.(2021)Zhang, Yao, and Deng]{zhang2021fast}
Juyong Zhang, Yuxin Yao, and Bailin Deng.
\newblock Fast and robust iterative closest point.
\newblock \emph{IEEE Transactions on Pattern Analysis and Machine Intelligence}, 44\penalty0 (7):\penalty0 3450--3466, 2021.

\bibitem[Zhang et~al.(2023)Zhang, Yang, Zhang, and Zhang]{zhang20233d}
Xiyu Zhang, Jiaqi Yang, Shikun Zhang, and Yanning Zhang.
\newblock 3d registration with maximal cliques.
\newblock In \emph{Proceedings of the IEEE/CVF Conference on Computer Vision and Pattern Recognition}, pages 17745--17754, 2023.

\end{thebibliography}
}


\end{document}